%% file: main.tex
\documentclass[sigconf, nonacm]{acmart}

\usepackage{graphicx}
\usepackage{algorithm}
\usepackage{algpseudocode}
\usepackage{amsmath}
\usepackage[font=small,labelfont=bf]{caption}
\usepackage{subcaption}
\usepackage[utf8]{inputenc}
\usepackage{amsmath}
\usepackage{amsthm}
\usepackage{scalerel}
\usepackage{bbm}
\usepackage{algorithm,algpseudocode}

\usepackage{acro}
\usepackage{graphicx}
\usepackage{tikz}

\usepackage{multirow}
\usepackage{enumitem}
\usepackage{pgfplots}
\usepackage{pgfplotstable}
\usepackage{amsfonts}
\usepackage{amsmath,amstext}
\usepackage{textcomp}
\usepackage{comment}
\usepackage{gensymb}
\usepackage{balance}
\pgfplotsset{compat=newest}
\usepgfplotslibrary{groupplots,dateplot}
\usepackage{pgfplotstable}
\usepgfplotslibrary{statistics}
\usepgfplotslibrary{patchplots}
\usepackage[capitalise]{cleveref}
\usetikzlibrary{pgfplots.statistics, pgfplots.colorbrewer,pgfplots.dateplot,
        shadows,
        matrix} 
\usepackage{array}

\usepackage{filecontents}
\usepackage{csvsimple}
\usepackage{pgfplots} % loads tikz
\usepackage{newfloat}
\usepackage{listings}
\DeclareCaptionStyle{ruled}{labelfont=normalfont,labelsep=colon,strut=off} % DO NOT CHANGE THIS
\floatstyle{ruled}
\newfloat{listing}{tb}{lst}{}
\floatname{listing}{Listing}
\DeclareAcronym{ac}{
  short = AC,
  long  = Assurance Case,
}

\DeclareAcronym{ml}{
  short = ML,
  long  = machine learning,
}

\DeclareAcronym{bo}{
  short = BO,
  long  = Bayesian Optimization,
}

\DeclareAcronym{dnn}{
  short = DNN,
  long  = deep neural network,
}

\DeclareAcronym{cps}{
  short = CPS,
  long = cyber-physical systems,
}

\DeclareAcronym{sdl}{
  short = SDL,
  long = scenario description language
}

\DeclareAcronym{av}{
  short = AV,
  long = autonomous vehicle
}

\DeclareAcronym{lbc}{
  short = LBC,
  long  = learning by cheating,
}

\DeclareAcronym{lec}{
  short = LEC,
  long = learning enabled component,
}

\DeclareAcronym{knn}{
  short = K-NN,
  long = k-nearest neighbors,
}

\DeclareAcronym{mlp}{
  short = MLP,
  long = multilayer perceptron,
}

\DeclareAcronym{ood}{
  short = OOD,
  long = out-of-distribution
}

\DeclareAcronym{gps}{
  short = GPS,
  long = global positioning system
}
\DeclareAcronym{imu}{
  short = IMU,
  long = inertial measurement unit
}

\DeclareAcronym{rl}{
  short = RL,
  long = reinforcement learning
}

\DeclareAcronym{ucb}{
  short = UCB,
  long = upper confidence bound
}

\DeclareAcronym{lut}{
  short = LUT,
  long = lookup table
}

\DeclareAcronym{mdp}{
  short = MDP,
  long = markov decision process
}

\DeclareAcronym{smdp}{
  short = SMDP,
  long = Semi-Markov Decision Process
}

\DeclareAcronym{gp}{
  short = GP,
  long = Gaussian process
}

\DeclareAcronym{mcts}{
  short = MCTS,
  long = Monte Carlo tree search
}

\DeclareAcronym{ap}{
  short = AP,
  long = autopilot
}

\DeclareAcronym{ds}{
  short = DS,
  long = dynamic simplex strategy
}

\newcounter{reviewercomment}[section]

\definecolor{darkblue}{rgb}{0,0,0.5}

\title{Dynamic Simplex: Balancing Safety and Performance in Autonomous Cyber Physical Systems}
\author{Baiting Luo}
\affiliation{%
  \institution{Vanderbilt University}
  \country{USA}}
\email{baiting.luo@vanderbilt.edu}

\author{Shreyas Ramakrishna}
\affiliation{%
  \institution{Vanderbilt University}
  \country{USA}}
\email{shreyas.ramakrishna@vanderbilt.edu}

\author{Ava Pettet}
\affiliation{%
  \institution{Vanderbilt University}
  \country{USA}}
\email{ava.pettet@vanderbilt.edu}

\author{Christopher Kuhn}
\affiliation{%
  \institution{Technical University of Munich}
  \country{Germany}}
\email{christopher.kuhn@tum.de}

\author{Gabor Karsai}
\affiliation{%
  \institution{Vanderbilt University}
  \country{USA}}
\email{gabor.karsai@vanderbilt.edu}

\author{Ayan Mukhopadhyay}
\affiliation{%
  \institution{Vanderbilt University}
  \country{USA}}
\email{ayan.mukhopadhyay@vanderbilt.edu}
\begin{document}

% % REMOVE THIS TO REVERT TO CAMERA READY
% \pagestyle{fancy}
% \rhead{}
% \cfoot{\thepage}

% %\setlength{\textfloatsep}{2pt}
% \sloppy

\input{sections/abstract}

\maketitle

\pagenumbering{arabic}
\input{sections/introduction}
\input{sections/problem}
\input{sections/approach_new}
\input{sections/evaluation}
\input{sections/related_work}
\input{sections/conclusion}
%\input{sections/appendix}

\balance
\bibliographystyle{ACM-Reference-Format} 
\bibliography{references.bib}
\clearpage
\newpage
\appendix

\input{sections/appendix}

\end{document}

%% file: sections/abstract.tex
\begin{abstract}
Learning Enabled Components (LEC) have greatly assisted cyber-physical systems in achieving higher levels of autonomy. However, LEC’s susceptibility to dynamic and uncertain operating conditions is a critical challenge for the safety of these systems. Redundant controller architectures have been widely adopted for safety assurance in such contexts. These architectures augment LEC “performant” controllers that are difficult to verify with “safety” controllers and the decision logic to switch between them. While these architectures ensure safety, we point out two limitations. First, they are trained offline to learn a conservative policy of \textit{always} selecting a controller that maintains the system's safety, which limits the system’s adaptability to dynamic and non-stationary environments. Second, they do not support reverse switching from the safety controller to the performant controller, even when the threat to safety is no longer present. To address these limitations, we propose a dynamic simplex strategy with an online controller switching logic that allows two-way switching. We consider switching as a sequential decision-making problem and model it as a semi-Markov decision process. We leverage a combination of a myopic selector using surrogate models (for the forward switch) and a non-myopic planner (for the reverse switch) to balance safety and performance. We evaluate this approach using an autonomous vehicle case study in the CARLA simulator using different driving conditions, locations, and component failures. We show that the proposed approach results in fewer collisions and higher performance than state-of-the-art alternatives.
% Our complete implementation is available online in an anonymous repository.
\end{abstract}

% \Ayan{I added line numbers and wide margins for reviewing. Do not remove for now. We will remove it before submission.}

%% file: sections/introduction.tex
\section{Introduction}
\label{sec:introduction}
Autonomous \ac{cps} are an important component of many applications in the fields of medicine, aviation, and the automotive industry. Such systems are often equipped with LEC that are trained using \ac{ml} methods~\cite{dartmann2019big}. 
A critical challenge for these systems is making safe and efficient decisions under unanticipated system faults and dynamically changing operating conditions~\cite{varshney2017safety}. However, recent incidents involving \acp{av} from automotive companies such as Waymo, Tesla, Uber, and Cruise~\cite{DMV} illustrate the complexity of this decision-making process. Indeed, the National Highway Traffic Safety Administration (NHTSA) released a summary report that highlighted 392 crashes involving \acp{av} in the United States between June 2021 and May 2022~\cite{nhtsa}.

% , this is hard \ad{find newer cases. look at what happened with cruise.}

%While autonomous \ac{cps} have illustrated exceptional performance in a variety of challenging problems, recent accidents such as Tesla's autopilot crashes~\cite{vlasic2016self} and the fatal Uber self-driving car accident~\cite{kohli2019enabling} demonstrate that these systems can still fail, often with devastating consequences. 

A common mechanism for dealing with failures and ensuring safety, especially in \ac{cps}, is the usage of \textit{controller-redundant} architectures, e.g., the simplex architecture~\cite{seto1998simplex}
%One commonly used approach for providing (some) safety assurance to \ac{cps} are \textit{controller-redundant} architectures, e.g., the simplex architecture~\cite{seto1998simplex} 
and controller sandboxing~\cite{bak2011sandboxing}. Such architectures typically augment \ac{cps} that use a high-performing but unverifiable controller (the \textit{performant} controller) with a verified controller (the \textit{safety} controller)~\cite{bak2014real}. A decision logic, often uses a verification-based approach trained with a safety-based utility function or a simple set of domain rules, triggers a switch from the performant controller to the safety controller under unsafe operating conditions or system faults. Verification-based approaches like linear matrix inequality~\cite{seto1999case}, reachability analysis~\cite{bak2011sandboxing}, and safety certificates~\cite{prajna2004safety} have also been explored for this decision logic. Such techniques have been widely (and successfully) used in practice, e.g., unmanned aerial vehicles~\cite{vivekanandan2016simplex}, remote-controlled cars~\cite{crenshaw2007simplex}, and industrial infrastructures~\cite{DBLP:conf/hicons/MohanBBYSC13}.

% Verification-based variants \ad{I dont understand the argument mixing verification based approaches and simplex approaches. We probably dont need this. I will revise this tomorrow.} \Shre{Abhishek, the main idea here is that most existing decision logic for Simplex are based on verification approaches of LMI or Reachability Analysis.} like linear matrix inequality~\cite{seto1999case}, reachability analysis~\cite{bak2011sandboxing}, and safety certificates~\cite{prajna2004safety} are also widely used for safe decision making. 
While these approaches have shown promising results, there are two major limitations. \underline{First}, the decision logic is generally trained offline. While offline training provides the advantage of invoking the policy almost instantaneously when making decisions, such policies can often become stale in non-stationary and dynamic conditions~\cite{hoel2019combining, pettet2022decision}. \underline{Second}, the decision logic is usually designed to \textit{only perform a one-way switch}, i.e., when the system under consideration detects an imminent threat to safety, the logic dictates a switch from the performant controller to the safety controller. Once such a switch is made, the control remains in the safety mode forever (barring some exceptions that perform the reverse switch based on system stability~\cite{desai2019soter,johnson2016real}). However, once the threat no longer exists, using the safety controller could delay or ignore the system's mission-critical objectives~\cite{johnson2016real}. 

However, the reverse switch 
% \ava{If there's time, I suggest different labels for the switches -- I think that ``forward'' and ``reverse'' switching is not very descriptive. Maybe just ``P-S switch'' and ``S-P switch'' for ``Performance to Safety'' and ``Safety to Performance'' switching, respectively?}
, i.e., transitioning back to the performant controller from the safety controller, is highly non-trivial for several reasons. \underline{First}, in real-world \ac{cps}, safety is paramount, and a myopic switch could be detrimental to the overall health of the system and related entities, including humans. As a result, it is imperative that a careful and non-myopic evaluation is done on the evolution of the system and possible exogenous factors before switching back to the performant controller. \underline{Second}, these exogenous factors could be dynamic~\cite{pettet2021hierarchical}; such variation makes it necessary that the logic is equipped to perform planning with the most up-to-date information available at hand (e.g., through online planning). However, online approaches are slow (compared to their offline counterparts), which inhibits their usage in practice. 
% \ava{What does our architecture do to address the slow convergence of online approaches for reverse switching? If we don't address it, I don't think it should be listed as a challenge.}
\underline{Third}, although the reverse switch is crucial for improving the system's performance, frequent back-and-forth switching among the controllers can be detrimental to stability and performance. While these factors make designing the reverse switching logic challenging, one advantage compared to the forward switching logic is that the system's safety is not sensitive to the computation time, allowing us to explore non-myopic, principled, albeit more computationally intensive algorithmic methods.

In this paper, we present a principled hybrid approach to address the challenges of balancing the safety and performance objectives in autonomous \ac{cps}. We make the following contributions: \textbf{1)} We present the dynamic simplex strategy (DS) that is online and allows for two-way switching while avoiding frequent back-and-forth arbitration. \textbf{2)} We formulate the decision-making problem as a semi-Markov decision process~\cite{janssen2013semi}. While we perform the forward switch (i.e., from the performant to the safety mode) myopically to prioritize safety, the decision for reverse switching to the performant controller is performed in a non-myopic manner to find a promising action by using \ac{mcts}. We present a combination of online heuristic search and domain-based safety rules for switching. \textbf{3)} To monitor the need for switching, we use a set of runtime data-driven safety monitors that collectively indicate the system's imminent risk by keeping track of system faults and uncertainties in the operating environment. \textbf{4)} We evaluate the proposed approach extensively through multiple \acl{av} studies in simulated urban environments using the CARLA simulator~\cite{dosovitskiy2017carla} and demonstrate that the proposed approach leads to fewer infractions and higher performance than state-of-the-art alternatives.

% \textbf{Outline}: The rest of this paper is organized as follows. We begin by outlining the problem formulation in \cref{sec:ps}. Then, we present the proposed approach in \cref{sec:approach} and show experimental results in \cref{sec:evaluation}. Finally, we summarize related research in \cref{sec:rw} and present our conclusion in \cref{sec:conclusion}. 

%% file: sections/problem.tex
\section{Problem Setup and Model}
\label{sec:ps}

\subsection{Problem Setup} Consider an autonomous \ac{cps} with both performance and safety objectives. We use the example of an autonomous vehicle for this paper. %The goal of a decision-maker, i.e., the system designer \ava{This is confusing, since `decision maker' also refers to the switching decision logic. Just call this system designer?}, 
The goal of the system designer is to enable control logic that determines operational parameters such as speed and steering angle. Instead of directly \textit{acting} on the parameters, such systems are equipped with controller(s) that affect operational parameters. The decision-maker must therefore select a controller, which in turn, selects the parameters. Typically, a performant controller, $C^p$, is used to ensure that the autonomous CPS focuses on its performance objectives (e.g., minimize the travel time)~\cite{gohari2020blending}. Performant controllers are often designed using \ac{ml}-based approaches and trained on data that closely mimics the operating conditions. Such data typically involves information from sensors like cameras, radar, and lidar to compute high-level trajectories or low-level control actions for the system. Formally, we denote a data point representing an operating condition as a \textit{scene}. For example, a scene could be a collection of scalar values denoting precipitation and the location of the vehicle over a few seconds. We assume that each scene is associated with a set of features $w \in \mathbb{R}^m$ that includes structural (spatial) features $w^s$ (e.g., the type of road, road curvature, and the presence of road signs) and temporal features $w^q$ (e.g., weather conditions) characterizing the operating conditions.\footnote{while we define features for autonomous vehicles, such attributes can capture arbitrary operating conditions relevant to any autonomous CPS.} Typically, the performant controller is trained on data from a large number of such scenes. 

%which is a short time trajectory of the system in the operating environment.

In trying to achieve the system's performance objectives, the performant controller may neglect the safety objectives~\cite{johnson2016real}. As a result, these systems are also equipped with several runtime monitors, a safety controller $C^s$, and a decision logic for safety assurance. The monitors raise alarms based on identifying different operational hazards (e.g., out-of-distribution (OOD) data for LEC) and system hazards (e.g., sensor failure), which can be either critical or non-critical. When a hazard is detected, the decision logic switches the system's control from the performant controller to the safety controller, which focuses exclusively on ensuring safety, e.g., the safety controller might reduce the system's speed, intervene through braking, or alert the driver for manual intervention. Given this setting, our goal is to design an approach that balances the safety and performance objectives of the CPS.

\subsection{Problem Formulation}
\label{sec:pf}
We refer to the autonomous CPS and the environmental conditions (i.e., $w$) as our \textit{system} of interest. We begin with the assumption that the decision-maker knows the spatial features $w^s$ \textit{a priori} for the finite set of scenes the vehicle will travel through. We assume that the future temporal features $w^q$ are unknown to the vehicle. In practice, information pertaining to the spatial features such as the curvature of the road can be retrieved easily. Note that this assumption is not critical (or important) for our formulation or solution approach; it is merely based on our domain of interest. 
%\ava{I don't think it is clear what your `domain of interest' is, and why it is relevant to this assumption.}. 

We consider the evolution of the system in continuous time. The dynamics of decision-making are governed by the following events: (a) when the spatial parameters $w^s$ change (e.g., the vehicle enters a new stretch on the road with curvature), (b) when the temporal parameters $w^q$ change (e.g., the weather changes), (c) when a component of the system fails, (d) when the traffic density changes, or (e) when the runtime monitor state changes. When an event occurs, the decision-maker must take an action, i.e., choose between operating in the performant mode or the safety mode. Note that the time between the events is governed by some exogenous distribution that is not necessarily memoryless; for example, the change of the spatial parameters depends on the speed of the car. To capture the non-memoryless transitions and the continuous-time evolution of the system, we model our decision-making problem as a semi-Markov decision process (SMDP)~\cite{janssen2013semi}. 

An SMDP can be represented by a tuple $\{ \mathcal{S},\mathcal{A},T,R,\tau \}$, where $\mathcal{S}$ is a finite set of states, $A$ is a finite set of actions that can be performed in a state, $T$ is the state-action transition model, $R$ is the reward function, and $\tau$ is the temporal distribution over state transitions. We describe each component of the SMDP below.

% Next, though the state space of the decision-making problem evolves in continuous time, for convenience we view it as a finite set of decision making states that evolve in discrete times. To explain this, consider a CPS traveling through one scene in a given route. During its journey in the scene, the system changes the state of the world, but presents no requirement for decision-making unless a \textit{decision event} occurs. We have three events that present scope for decision-making: (a) when the system enters a new scene, (b) when the weather changes or (c) when a component of the system fails. 

% We finally assume no two events can occur simultaneously. 

% Next, a key component of our decision-making problem is that the system physically travels across scenes, which makes the temporal transitions between decision-making states non-memoryless. This further makes the underlying stochastic process governing the system's evolution to be semi-Markovian~\cite{pettet2021hierarchical}. Therefore, we model the sequential decision-making problem of selecting an optimal controller as a \ac{smdp}, which can be represented as a tuple $\langle S,A,T,R,\tau \rangle$, where $S$ is a finite state space, $A$ is a set of actions that can be performed from a state, $T$ is the state-action transition model, $R$ is a reward function that decides the instantaneous reward for taking an action in a state, and $\tau$ is the temporal distribution over state transitions.

\textbf{State} We denote the finite set of states by $\mathcal{S}$. 
% We use $s_t \in \mathcal{S}$ to denote the state of the system at an arbitrary time $t$, which includes information about the scene, the controller driving the system, current component faults, and a counter to keep track of the number of controller switches so far (we keep track of the number of switches to penalize frequent switching). 
Formally, we represent the state $s_t \in \mathcal{S}$ by the tuple $(v_t,w^s_t,w^q_t,d_t,C_t,\Phi_t,\Psi_t, \omega_t)$, where $v_t$ is the velocity of the vehicle, $w^s_t$ and $w^q_t$ are the structural and temporal scene features, $d_t$ is the traffic density, $C_t \in \{C^p,C^s\}$ is the controller driving the system, $\Phi_t$ = $\{\phi^1_t, \dots , \phi^n_t\}$ is the failure state of the $n$ components (e.g., sensors), $\Psi_t$ is the runtime monitor state (e.g., OOD detector), and $\omega_t$ is a counter that keeps track of the number of switches that have been performed until the current time, with all of the variables being observed at time $t$. We assume that $\phi^i_t \in \{0,1\} \, \forall i,t$.

\textbf{Actions} We denote the set of all actions by $\mathcal{A}$. An action in our setting is restricted to selecting a controller. In this paper, we restrict our attention to two controllers---a safety controller and a performant controller. Our action space is therefore simplified to the binary choice of whether or not to switch the controllers. In principle, our problem formulation (and the solution approach) can accommodate an arbitrary number of controllers as part of the action space.

\textbf{Transitions}: The evolution of our system model is governed by several stochastic processes. First, the spatial parameter is governed by the track on which the system operates (known \textit{a priori}) and the system's speed, which in turn is a function of the controller in use. Second, the weather conditions, traffic density, sensor failures and runtime monitor states are governed by exogenous distributions. As our solution approach is based on exploring possible trajectories under the effect of the actions, we only need access to a set of generative models for simulating the transitions~\cite{pettet2021hierarchical}. We describe the specific models we use for such parameters in the evaluation section. The only deterministic update to a state under an action is that of the counter $\omega$, which is incremented by 1 every time the decision logic performs a switch between the controllers.

\textbf{Reward Function}: Rewards in an SMDP consist of a lump sum immediate reward upon taking an action and/or a continuous-time reward as the system evolves~\cite{janssen2013semi}. We model reward as the  sum of immediate rewards that capture both performance and safety objectives. Formally, the reward for an action $a \in \mathcal{A}$ in state $s_t \in \mathcal{S}$ includes a performance score $\lambda^p(s_t,a)$ and a safety score $\lambda^f(s_t,a)$. In our implementation, we model $\lambda^p$ as the chosen controller's (i.e., the action $a$'s) average speed and $\lambda^f$ as the controller's likelihood of collision given the state $s_t$. We estimate both terms by using surrogate models trained through historical data (we describe the exact estimation process in \cref{sec:surrogate}). The instantaneous reward is calculated as the weighted sum of the two scores:
\begin{equation}
R(s_t,a) = \alpha_1 \cdot \lambda^p(s_t,a) - \alpha_2 \cdot \lambda^f(s_t,a)
\label{eqn:fwd}
\end{equation}
where $\alpha_1$ and $\alpha_2$ are hyperparameters.

While such a function is sufficient to capture safety and performance objectives, practical constraints require that we prevent frequent back-and-forth switching among the controllers. % Essentially, switching to the safety mode can be done myopically to avoid an imminent hazard. However, switching from the safety to the performant mode must be non-myopic. 
Therefore, we include a third term called cost of switching $\lambda^c$ that adds a penalty based on the number of previously performed switches leading up to the current state. For an arbitrary state $s_t$, we use $\omega_t$ to compute this penalty (recall that $\omega_t$ tracks the number of controller switches). Specifically, $\lambda^c$ = 0, if the $\omega_t$ variable is 0 or 1; otherwise $\lambda^c$ = $\omega_t/m_s$, in which $m_s$ is the maximum number of switches that can happen during the planning horizon (i.e. future scenes considered during panning). However, we point out that the cost of switching frequently should not be used to limit the forward switch (from the performant to the safety mode) as it can compromise the safety of the autonomous CPS. Therefore, we calculate the reward for the forward switch according to \cref{eqn:fwd} and use the penalty term only for computing rewards for the reverse switch as shown below:
\begin{equation}
R(s_t,a) = \alpha_1 \cdot \lambda^p(s_t,a) - \alpha_2 \cdot \lambda^f(s_t,a) - \alpha_3 \cdot \lambda^c(s_t, a)
\label{eqn:rev}
\end{equation}
Based on the above SMDP, given a state, our goal is to choose actions based on a utility function (e.g., expected discounted reward). We describe the exact criteria and our approach below.

% \ad{this should have a reference to papers and forward section where you show how this can be calculated. Then refer to this in the mCTS section}

%% file: sections/approach_new.tex
\section{Dynamic Simplex Strategy}
\label{sec:approach}
\begin{figure}[t]
 \centering
 \includegraphics[width=\columnwidth]{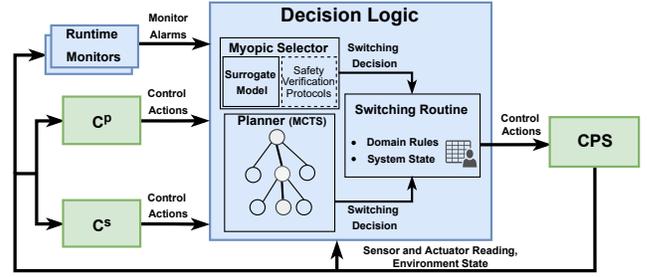}
 \caption{Overview of the proposed dynamic simplex strategy. The blocks in blue are designed through the solution approach. The green blocks represent generic autonomous \ac{cps} components and controllers.}
 \label{fig:overview}
\end{figure}

A schematic diagram of our approach is shown in \cref{fig:overview}. We switch between the controllers based on the following criteria: for the forward switch (i.e., from the performant to the safety controller), we take the action that maximizes the myopic one-step reward (based on \cref{eqn:fwd}), which ensures that any imminent threats to safety are thwarted based on historical data. Furthermore, our method assumes some safety verification protocols will be given to provide conditions for switching into the safety controller (e.g., conditions provided by the proof of safety for the safety controller) when it is applied to real-world applications. Naturally, the decision-maker cannot be entirely myopic about action selection; the performance score $\lambda^p$ and the safety score $\lambda^f$ capture some non-myopic effects of taking an action by leveraging a surrogate model trained using historical data. However, note that the reverse switch occurs \textit{after} the control logic had previously decided to switch to the safety controller; this decision must have resulted from an imminent threat to the safety of the CPS. Hence, to ensure that the CPS can safely switch back to the performant mode, we do non-myopic planning and take the action that maximizes the expected discounted cumulative reward. 
These criteria essentially form the core of our dynamic simplex strategy. To actuate the strategy, we use the following components: 1) %a myopic action selector that uses a non-parametric supervised learning approach based on historical data to perform the forward switch
a myopic action selector that uses given safety verification protocols and a complementary neural network-based surrogate model trained with historical data and ; 2) a non-myopic planner based on an approximate heuristic search algorithm to perform the reverse switch; and 3) a set of runtime monitors to monitor changes in environmental parameters and sensor faults. 

%\begin{figure}[t]
% \centering
% \includegraphics[width=\columnwidth]{figures/workflow_new1.pdf}
% \caption{Overview of the proposed dynamic simplex strategy. The blocks in blue are designed through the solution approach. The green blocks represent generic autonomous \ac{cps} components and controllers.}
% \label{fig:overview}
%\end{figure}

We describe the proposed strategy briefly here. When an event occurs, the control logic first checks which controller is driving the system. If the performant controller is operating the system, the switcher activates the myopic action selector based on the trained surrogate model (or a verification protocol) to get the performance and safety scores of both controllers. No switching is performed if the performant controller has a higher reward. Otherwise, a forward switch is performed. Next, if the safety controller is operating, the switcher activates the planner, which uses a set of generative models to simulate possible future trajectories and picks an action that maximizes the expected discounted cumulative reward. To balance exploration and exploitation in the future trajectories, we use \ac{mcts}. In our setting, if a new event occurs while a decision is being computed, the computation is immediately terminated, and decision-making is initiated with the new state. We describe each decision-making component below.

% Before we explain each component, we present a controller transition diagram (\cref{fig:example-transition}) to illustrate decision-making in the proposed framework. Events that trigger the need to make decisions are highlighted in red. 

\subsection{Myopic Action Selector}
\label{sec:surrogate}

The role of the myopic action selector is to decide whether to switch from the performant controller to the safety controller, i.e., the forward switch. A number of varied works have proposed different potential methods to trigger the decision logic, e.g., safety verification methods such as reachability analysis, which computes a set of states reachable within some number of time steps and then checks if this reachable set contains states outside the system's safe region~\cite{phan2020neural}; and OOD detection methods, which detect the data that are not similar to the
data used for training~\cite{DBLP:conf/iccps/YangKDL22}.

However, the existing safety verification methods are subject to limited efficiency and accuracy for
safety-critical systems that are time-sensitive and operate in dynamic environments~\cite{DBLP:conf/iccps/LiuHWZZ22}. Further, computing the exact reachable set for most nonlinear systems is a complex problem~\cite{DBLP:journals/jcss/HenzingerKPV98}, and deployment of LEC aggravates the complexity of the problem. However, safety verification methods for neural network control systems are important components to ensure the safety of complex CPS. The myopic selector we propose (\cref{fig:overview}) can easily accommodate outputs of safety verification protocols and complement arbitrary surrogate models into decision-making procedure. This integration is orthogonal to the main contribution we make; in this paper, we show how a computationally cheap and safety-focussed forward switch can be complemented with a data-driven non-myopic reverse switch to balance safety and performance.

More specifically, the goal of the model used for the forward switch is to take as input the current state $s_t$ and return an aggregated score that captures the immediate threat to the safety of the system, coupled with some performance objective. In this work, the score is composed of the collision likelihood of both the controllers and a measure of performance based on normalized average speed (see Eq. (1)). Specifically, we train a \ac{dnn} using historical data to estimate these parameters to show the proposed technique works well by only using a surrogate model. Specifically, we use historical data of the performant and safety controllers pertaining to operations in similar environments. This data is curated from a large number of prior simulations, and the simulated observations are stored in a tabular format (we describe the exact curation process in \cref{sec:setup}). Furthermore, the safety of the system can be further guaranteed as long as a safety verification protocol is given. Finally, the myopic action selector chooses the best action based on \cref{eqn:fwd}.

\subsection{Non-Myopic Planner}

The decision to switch from the safety controller to the performant controller is more involved than the forward switch, and has not been addressed in prior work, partly due to its complexity. Given a state $s_t$ such that $C_t$ = $C^s$, we use \ac{mcts} to evaluate whether or not to switch the controller to $C^p$. \ac{mcts} is a heuristic search algorithm for sequential decision-making~\cite{browne2012survey}. It enables online planning in a way that can incorporate changes in environmental parameters at decision-time; also, it is an anytime algorithm, which enables the system-designer to constrain the computation time based on domain-specific requirements. 

% We choose \ac{mcts} as the flexibility to perform online planning ensures that any changes in the environmental parameters can be incorporated directly at decision-time. Also, \ac{mcts} is an anytime algorithm, which enables the decision-maker to put constraints on the computational time and latency that can be afforded to make a decision depending on domain-specific requirements. 

\ac{mcts} operates by iteratively building a search tree that represents future trajectories. The tree nodes represent the states, and the edges represent the actions that mark the state transitions. The search begins with a root node that denotes the current state. In each simulation, a child node is recursively selected until a leaf node is reached. Unless this leaf node denotes the end of the planning horizon, an action is taken in this state node, and the tree is expanded. To estimate the value of an action from a node, the algorithm simulates a ``rollout" from the child node to the end of the planning horizon with a computationally cheap default policy. Using the algorithm requires three components: (1) generative models to simulate the future states, (2) a tree policy to navigate the tree search, and (3) a default policy to estimate the value of an action. We describe these components below.

% \adi{write here a paragraph that says surrogate models -- describe that they identify the likelihood of incidents and performance of the controller in a given scene. Explain that you need a function like that. Say that function can be trained in multiple ways -- point to anti carla paper. in experiments describe how you trained it}

% The tree can be explored asymmetrically to search for the actions with the most promising rewards.

\textbf{Generative Models}: We use a set of generative models to sample future trajectories. First, we use a model for sampling future weather parameters, conditioned on the current weather. We learn the weather model using the historical data gathered from simulations. % We describe the exact parameters in the evaluation section. 
Second, we use a model for sampling average traffic density conditioned on current traffic density, also learned based on simulated data.
Third, we model sensor failures by gathering historical data about the sensors and their susceptibility to adverse weather conditions and lighting levels. While some of these failures (e.g., bright images) are rectified over time, other failures (e.g., broken lens) persist until replacement. 
Fourth, we model the runtime monitor alarms by learning the duration and the arrival time of such states using historical data. Finally, we use the surrogate model and any given safety verification protocol discussed in the previous section to identify the safety score and the performance score of the controllers. While we use different distributions and neural networks for learning the generative models, the proposed framework is agnostic to the model used. For the sake of brevity, we present a detailed description of the generative models
in~\cref{appendix:generative_model}.

% For example, a typical digital camera's failure rate is roughly 0.1\% over $2$ years of operation.

% \Shre{I've referred to the surrogate model here. It is our third model to give the collision likelihood and performance}  

%\ad{we need to also write how we calculate the state update. that is the next step of the tree -- in other worlds we need to describe the rollout policy}

\textbf{Tree Policy}: We perform action selection within the tree using the standard Upper Confidence bound for Trees (UCT) algorithm~\cite{KocsisS06}.
% , which selects the action that maximizes the following score for a node representing state $s_t$ in the tree:
% \[
%     UCB(s_t,a) = Q(s_t,a) + c_{uct} \sqrt{\frac{\ln{N(s_t)}}{N(s_t,a)}}
% \]
% where $Q(s_t,a)$ is the current estimated value for taking the action $a$ in $s_t$, $N(s_t,a)$ denotes how many times $a$ has been taken in $s_t$, $N(s_t)$ is the total number of times $s_t$ has been visited, and $c_{uct}$ is a hyperparameter that trades off between exploitation and exploration---a lower value emphasises the current estimated value of an action $Q(s_t,a)$ (exploitation) while a higher value guides the search towards actions that have been taken less frequently (exploration). 

% higher value for $c_{uct}$ puts less emphasis on the current value estimation $Q(s_t,a)$ and guides search towards actions taken less frequently,  The first term $Q(s_t,a)$ is the exploitation term that biases the search toward the nodes which appear promising. The second term is the exploration term which encourages attempts at the under-represented action for the current state.

\textbf{Default Policy}: To simulate rollouts, it is common to use a computationally cheap default policy~\cite{pettet2021hierarchical}. Our default policy is random, i.e., we randomly select between making a switch or staying with the current controller in use. 

% an action that should be taken when the safety controller is operating the system. If the chosen action is ``to switch" to the performant controller, we will sample the future state only from a set of states with the driving controller as the performant controller. The idea behind biasing the sampling process is that we want to see how the expected rewards will be if the system continues using the performant controller in the future.

\textbf{Tree Search}: Each search begins by initializing the current state as the tree's root node. The core idea behind the algorithm is that the search tree over possible future trajectories is explored asymmetrically and iteratively, with the search being biased toward promising action trajectories. During each simulation, the tree policy (explained above) is used to select which leaf node is expanded, after which the default rollout policy quickly estimates the new node's value. 

\textbf{State Transitions}: Here, we describe how we perform the state transitions inside the tree. Although the temporal parameters (e.g., weather) evolve in continuous time, we discretize time and assume that such parameters are queried every $\tau^{q}$ units of time, e.g., our system can query the current weather every 20 seconds. In principle, each parameter can be queried at different frequencies depending on the domain of interest; in such a case, let $\tau^{q}$ denote the discrete time period after which the parameter with the highest frequency is queried. Each state transition caused by other events will evolve $t_q$ towards $\tau^{q}$, which denotes the transition of temporal parameters in our case. Once new temporal parameters are sampled, $t_q$ is reset to $\tau^{q}$.

At a given state $s_t$ (at time $t)$, the state of the system consists of the location, which can be defined by the structural parameters of the scene (i.e., $w^s_t$) and the velocity of the vehicle $v_t$. Depending on the action taken, $v_t$ can change. It is then trivial to compute the time taken by the vehicle to the next location (i.e., structural component of scene) as we assume that the structural features are known \textit{a priori} (see \cref{sec:ps}). We denote this arrival time by $t_e$. The time $t_e$ can then be compared with $\tau^{q}$ to populate the structural features of the next state, i.e., determine if the vehicle has moved to the next location. The other temporal parameters, the future average traffic density $d$ across a scene, the arrival time of sensor failures, and the arrival time of the next runtime monitor state are sampled according to the generative models. The earliest arriving event leads to the transition of the state.
Finally, we compute the reward for a state-action pair by querying a surrogate model (denoted by $G$), which is also used to retrieve the velocity of the vehicle inside the tree (for future states given an action). We present the complete search algorithm algorithm in \cref{algo:mcts}.

\input{algorithms/mcts}
%\ad{where is the reward function specified and where do you explain why the reward function shows up}

%\ad{I see the tree policy -- but not how the next actions are sampled and the next new state is determined}

%\ad{add how the rollout occurs here}

%\adi{Ideally by this time you should have clearly identified some parameters -- the slow speed for controller change, the tree depth, the time of change of weather, the ratios in the reward function.. by the way I do not see the reward function here. explain it with the tree search and rollout.}

\subsection{Runtime Monitors}
% \textcolor{red}{Shre: I added this paragraph to talk about faults and failures}
To ensure the safety of the autonomous CPS, it is imperative that we keep track of the relevant state and environmental parameters. It is especially essential to track sensor faults which can lead to catastrophic accidents~\cite{Resonate}. Several automotive standards, such as the ISO 26262~\cite{26262}, categorize faults as: (a) permanent faults, which persist until removed or repaired (e.g., a broken camera), and (b) intermittent faults, which eventually go away on their own (e.g., occlusion). Both types of faults must be monitored and incorporated into decision-making. 
% While permanent faults are relatively easy to detect because they fail the sensor at a specific failure rate, intermittent faults are regarded as the most challenging fault to detect and diagnose. These faults increase the operational risk of the system, thereby impacting its functional safety. Therefore, it is crucial to identify them, associate them with different risk levels, and design different safety mechanisms (e.g., anomaly detectors) to prevent, detect or mitigate their occurrence.  

In this paper, we deal with perception-based autonomous \ac{cps}. Therefore, we consider faults in the three cameras used in the autonomous vehicle in our case study (discussed in the experiments section). While many types of faults can be associated with a digital camera, we focus on the common fault of occlusion~\cite{ceccarelli2022rgb}. We train a monitor using prior data and only use inference on the trained models at decision time. 
% We design a blur detector that leverages the variance of the Laplacian operator across an image~\cite{pech2000diatom} to detect if the image is blurred. The operator highlights the image pixels with large changes in intensity. A variance in the operator's score below an exogenous threshold is used to detect blurred images. 
Specifically, based on prior work for occlusion detection in \ac{av}~\cite{Resonate}, we train a model to detect continuous blobs of black image pixels. We mask an image to find connected black pixels in it and then color these pixels as white. Then, we calculate the percentage of white pixels in the image and use an exogenous threshold on the resulting value to detect occlusion. We describe the parameter values in the evaluation section~\ref{sec:setup}.

We also implement the real-time OOD detector introduced by \citeauthor{faiyangood}~\cite{faiyangood} as one of our runtime monitors. We train a variational autoencoder (VAE) and utilize the reconstruction error for anomaly detection. Given an input, the trained decoder is used to sample independent and identically distributed (IID) samples from the latent space; then, the reconstruction error is used as a nonconformity measure within inductive conformal anomaly detection (ICAD). Given a sample, if the p-value computed by ICAD is smaller than a threshold, this test sample is hypothesized to be an OOD example. Finally, the computed $p$-values are used to construct the martingale, which the stateful detector uses to classify an input as an OOD example.

% In our setting, we use two runtime monitors on the three cameras used in the autonomous vehicle we study. Specifically, we focus on occlusion and blurring of images, two common camera-related faults~\cite{ceccarelli2022rgb}. We train both monitors using prior data and only use inference on the trained models at decision time. We design the blur detector that leverages the variance of the Laplacian operator across an image~\cite{pech2000diatom} to detect if the image is blurred. The operator highlights the image pixels with large changes in intensity. A variance in the operator's score below an exogenous threshold is used to detect blurred images. 
% We also design an occlusion detector to detect continuous blobs of black image pixels. We mask an image to find connected black pixels in it and then color these pixels as white. Then, we calculate the percentage of white pixels in the image and use an exogenous threshold on the resulting value to detect occlusion. We describe the parameter values in the evaluation section.

% \adi{we need a section that describes a bit of situations that can be called sensor failures. why are they important and what happens when they are intermittent.}

\subsection{Switching Routine and Domain Rules}
% \textcolor{red}{Shre: I modified this paragraph with the comments}
Finally, we acknowledge that data-driven decision-making must be coupled with a switching routine that uses appropriate domain rules and the system's state information to ensure that the decision can be smoothly implemented. This is necessary because the performant and the safety controllers operate at different performance levels (both controllers have different achievable high speeds), and performing an instantaneous controller transition can impact the system's stability. For example, in case of a forward switch, the system might be operating at a higher speed with the performant controller, which the safety controller cannot handle. Therefore, instead of an instantaneous controller transition, we consider the system's physical state and domain-specific rules to determine a feasible transition. Note that a forward switch, i.e., switching to the safety mode, must be performed irrespective of other factors to ensure the system's safety. As a result, for the forward switch, we design the decision logic to trigger a change in the control action (e.g., decrease speed). However, in the case of the reverse switch, we design the decision logic to trigger a change in the control action as well as domain rules. Intuitively, we switch from the safety mode to the performant mode within specific areas of the feature space to ensure safe transitions. In our setting, we enable the switch on the main roads, overpasses, and freeways; however, we disable reverse switching on intersections, lane changes, and roundabouts. We present the ablation study of domain rules in~\cref{appendix:ablation study}.

%% file: algorithms/mcts.tex
\algrenewcommand\algorithmicrequire{\textbf{Input:}}
\algrenewcommand\algorithmicensure{\textbf{Output:}}
%\Parameter number of tree simulations $M_n$, parameter controlling tree exploration $c_{uct}$
\begin{algorithm}[hbt!]
    \footnotesize{}
    \caption{Monte Carlo Tree Search (MCTS)}
	%\textbf{Parameter}: number of tree simulations $M_n$, parameter controlling tree exploration $c_{uct}$\\
    \begin{algorithmic}[1]
    \Require current state $s_t$, generative models $M$,  surrogate model $G$, query time $\tau^{q}$, $t_q = \tau^{q}$, number of tree simulations $I_n$, parameter controlling tree exploration $c_{uct}$
    \Ensure Policy $\pi$
    \Function{MCTS}{$s_t$}
        \State initialize $s_t$ as th
We present the complete search algorithm algorithm in \cref{algo:mcts}.e root node
        \For{$m=1,\cdots,I_n$}
            \State \textbf{Tree\_Search}($s_t$)
    	\EndFor
    	\State $\pi(a|s_t) \longleftarrow \frac{N(s_t,a)}{N(s_t)}$ \Comment{switching policy}
    	\State \Return $\pi$
    \EndFunction \\
    
    \Function{Tree\_Search}{$s_t$} \Comment{recursive tree search}
    \If{$s$ is terminal}
        \State $r \longleftarrow R(s_t,a)$
        \State \Return $r$
    \ElsIf{$s$ not visited}
        \State $r \longleftarrow rollout(s_t)$ \Comment{rollout}
        \State \Return $r$
    \EndIf
    \State $a \longleftarrow \operatorname*{argmax}_{a \in A} UCB(s_t,a, c_{uct})$ 
    \State $\hat{v} \sim G(s_t,a)$ \Comment{sample new velocity}
    \State $\hat{t}_o \sim M(s_t,a)$ \Comment{sample the duration of $\Psi_t$}
    \State $t_e = \textbf{distance}(w^{s}_{t},\hat{v})$ \Comment{estimate the time to arrive in the next structural scene}
    \State $\hat{t}_f \sim M(t,min(t_e,\hat{t}_o,t_q))$ \Comment{sample if sensor failure will happen before any other event and its arrival time}
    \State
    \If {$\hat{t}_f$ exist}
        \State $s^*_{t=t+\hat{t}_f} \sim M(s_t)$ \Comment{Sample new $\Phi$ and $d$}
        \State $t_q \longleftarrow t_q - \hat{t}_f$
    \ElsIf {$\hat{t}_o$ < $t_e$ and $\hat{t}_o$ < $t_q$}
        \State $s^*_{t=t+\hat{t}_o} \sim M(s_t)$ \Comment{Sample new $\Psi$ and $d$}
        \State $t_q \longleftarrow t_q - \hat{t}_o$
    \ElsIf{$t_e < t_q$}
        \State $s^*_{t=t+t_e} \sim M(s_t)$ \Comment{Sample new $w^{s}$ and $d$}
        \State $t_q \longleftarrow t_q - t_e$
    \ElsIf{$t_e = t_q$}
        \State $s^*_{t=t+t_{q}} \sim M(s_t)$ \Comment{Sample new $w^{s}$, $w^{q}$, and $d$}
        \State $t_q \longleftarrow \tau^{q}$
    \Else
        \State $s^*_{t=t+t_{q}} \sim M(s_t)$ \Comment{Sample new $w^{q}$ and $d$}
        \State $t_q \longleftarrow \tau^{q}$
    \EndIf
    \State \textbf{save} $t_q$
    % \State $r \longleftarrow R(s,a,s^*)$
    \State $r \longleftarrow R(s_t,a) + \gamma Tree\_Search(s_t^*)$ \Comment{perform tree search}
    \State $N(s_t, a)$ $\longleftarrow$ $N(s_t,a)$ + 1
    \State $Q(s_t, a) \longleftarrow Q(s_t, a) + \frac{r-Q(s_t, a)}{N(s_t,a)}$
    \State $N(s_t)$ $\longleftarrow$ $N(s_t)$ + 1
    
    \State \textbf{return} $r$
        
    \EndFunction

	\end{algorithmic} 
	\normalsize{}
\label{algo:mcts} 
%\vspace{-0.05in}
\end{algorithm}

%% file: sections/evaluation.tex
\section{Experiments}\label{sec:evaluation}
% In this section, we present several experiments to evaluate the usability of the proposed dynamic simplex strategy. These experiments were set up using an \acl{av} example in the CARLA simulator \cite{dosovitskiy2017carla} inside a docker, which was set up to utilize one NVIDIA Titan XP GPU. The details of the experiments and the \ac{av} setup are discussed below.

% \subsection{Setup}

% \subsubsection{Simulation Setup}
%\begin{figure}[t]
% \centering
% \includegraphics[width=0.8\columnwidth]{figures/tracks1.pdf}
% \caption{Samples of the tracks as captured by the center camera attached to the \ac{av}. The occlusion in the track 4 image is because of a camera failure.}
% \label{fig:regions}
%\end{figure}

We evaluate the proposed dynamic simplex strategy on an \ac{av} example in CARLA simulation~\cite{dosovitskiy2017carla}. We create $10$ tracks in two urban areas (towns) of the simulator. \cref{fig:regions} in \cref{appendix:AV_setup} illustrates a snapshot of these tracks with different weather conditions.
% Ten waypoints define the geometry of the tracks. We divide each track into five road segments containing two waypoints each \Ayan{What are waypoints?}. 
In total, we have $50$ road segments with various structural scene features. We consider the road type, road curvature, presence of traffic signs, and traffic density as structural (spatial) features $(w^s)$. We consider cloudiness, precipitation, and precipitation deposit parameters as temporal features $(w^q)$. To vary weather, we randomly sample a change in the neighborhood of the current parameters to avoid sudden (drastic) fluctuations in weather. Our complete simulation setup and implementation is available online.\footnote{See: \url{https://anonymous.4open.science/r/DynamicSimplex-7D23}}

% We leverage the data generation framework from 
% introduced by ramakrishna \emph{et al.}~\cite{ramakrishna2022risk} to automatically generate the data required for the experiments.\Ayan{What is this data generation framework?} 

% Next, for testing the logic, we used $5$ tracks that included $3$ tracks used during training and $2$ new tracks, one each from town3 and town5. The new tracks had similar scene\_labels that matched one of the $10$ tracks used during training. \cref{fig:regions} illustrates a snapshot of a few of these regions from different weather conditions.

% For automating the experiments, we leverage the automated test case generation framework proposed by Ramakrishna \emph{et al.}~\cite{ramakrishna2022risk}. The framework uses a \ac{sdl} for describing the weather conditions and the state of the system (e.g., sensor and actuator faults) in a scene. The language is also interfaced with a random sampler to generate different values for the weather parameters and sensor faults from their respective distributions. 

\input{graphs/travel_safe_boxplot}

\subsection{Setup}
\label{sec:setup}
\textbf{\ac{av} setup}: We show the system block diagram of the \ac{av} in~\cref{appendix:AV_setup}. It is primarily driven by a high-performant \ac{ml}-based controller called as \ac{lbc}~\cite{chen2020learning}, which uses a \ac{dnn} for navigation. The controller uses six sensors, including three forward-looking cameras, an \ac{imu}, a \ac{gps}, and a speedometer. For the safety controller, the system uses an autopilot controller, which is the safety controller in our setup. The controller uses the \ac{gps}, the speedometer, and the semantic segmentation camera sensor to compute the control action. We describe the exact operation of both the controllers in~\cref{appendix:AV_setup}. All experiments were run on a desktop computer with AMD Ryzen Threadripper 16-Core
Processor, 4 NVIDIA Titan XP GPUs, and 128 GiB RAM. 
%Our implementation as well as the data used for the analysis are available as part of the supplementary material.

\textbf{Data for Surrogate Model}: We run $111800$ simulations using each controller in the $50$ road segments across the CARLA towns, which amounts to $164$ hours of driving, for generating data for the surrogate models. We randomly vary weather parameters, traffic density, and introduce camera faults (image blur and camera occlusion) for one or more of the available cameras in the simulations. 

\textbf{Baselines}: We evaluate the performance of \ac{ds} against the following baseline controller configurations: (1) \ac{lbc} controller~\cite{chen2020learning}, (2) \ac{ap} controller~\cite{dosovitskiy2017carla}, and (3) traditional simplex architecture~\cite{seto1998simplex} with an offline decision logic for forward switching ($SA$). The forward switching is performed on the controller's historical performances, and (4) traditional simplex architecture with reverse switching ($GS$). For simplex configurations, both forward and reverse switching are performed with a myopic policy that maximizes the one-step reward (based on \cref{eqn:fwd}). 
%\Ayan{We need references here.}

% We call it greedy because the logic will always drive the vehicle with the controller having a higher performance. 

\textbf{Runtime Monitors}:
% The events triggering the decision-makers are: (a) change in the locations, (b) change in weather, and (d) a camera failure.
We assume our system is equipped with runtime monitors to detect the changes in location and weather conditions (it is trivial to design such monitors in practice). For sensor failures, we design a monitor that can detect occlusion in the \ac{av} as described in \cref{sec:approach}. Specifically, if white pixels are larger than $10\%$ (we set the threshold based on cross-validation), we consider an image to be occluded, otherwise, we mark the camera to not have occlusion. We obtain an F1-score of $98\%$ on the validation images. For detecting OOD inputs, we implement a real-time OOD detector as described in \cref{sec:approach}. We use similar parameters as in prior work~\cite{faiyangood}; we use a VAE to generate $10$ new examples for each given input, the martingale is then computed with the sequence of $p$-values computed by ICAD given the $10$ new examples. The threshold for stateful detector is set to $100$ to detect when the martingale becomes consistently large and if the input is an OOD example.

\textbf{Hyperparameters}:
% First, for the switching routine, the switching only happens when the system's speed is below 2.5 m/s. Second, 
% For the \ac{knn} algorithm to select data points from historical data in the vicinity of a given state, we
%For the \ac{knn} algorithm used in the surrogate model, we vary the number of neighbors $k$ between $[3, 4, 5]$, and use a neighborhood radius of $1.0$. We find $k$ = 5 to be the best parameter for our experiments by using a subset of the data for validation. 
For parameters of MCTS, the tree depth is set according to the simulation duration needed to finish the next three structural scenes. 
%\ava{Not clear to me what this means. It sounds like you have no limit on the lookahead horizon, with each trajectory terminating once the track is complete?}.
%\textcolor{red}{Ayan: Update the number of MCTS iterations and the parameter for lookahead.}
% Then, we estimate the time that the system may take to reach the future locations characterized by transitions in structural scene features with the mean velocities returned by the surrogate model; furthermore, we assume the mean sojourn time for temporal scene features are lower bounded to a variable $\tau_{w^t}$=$20$ seconds because the weather conditions do not normally change rapidly. Given the setup, even though $M_d$ is nondeterministic, it is upper bounded by $t_{\tau_{w^s}}/\tau_{w^t}+\tau_{w^s}$, in which $t_{\tau_{w^s}}$ is the time that the system approximately needs to take to finish travelling the next $\tau_{w^s}$ structural scenes.
We set the number of MCTS simulations per decision to 500, the exploration parameter $c_{uct}$ to $\sqrt{2}$, and the discount factor $\gamma$ to $0.9$. Finally, for the reward calculations, we choose the weights $\alpha_1$ = 1, $\alpha_2$ = 1, and $\alpha_3$ = 0.5 based on manual tuning. For tuning the hyperparameter, we use a subset of the data for validation, fix $\alpha_2$ at 1 (the weight on safety), and then vary the other parameters. We select the best weights based on a combination of travel times and safety score (we define these metrics below).
% We set the number of MCTS iterations to 1000, the exploration parameter $c_{uct}$ to $\sqrt{2}$, and the discount factor $\gamma$ to $0.9$. Finally, for the reward calculations, we choose the weights $\alpha_1$ = 1, $\alpha_2$ = 1, and $\alpha_3$ = 0.3 based on manual tuning. During tuning, we use a subset of the data for validation and keep $\alpha_1$ fixed at 1 (the weight on safety) and varied the other parameters. We selected the best weights based on a combination of travel times and safety score (we define these metrics below).

%\adi{make sure these parameters are explained in the approach. Also ensure that you atleast write how the novelty detector and collision likelihood detectors can be used within the system. }
\subsection{Results}

We run each controller $30$ times around each track. For each run, we start the initial scene with a random weather condition. We begin by evaluating the controllers without sensor failures. Then, we inject failures at random and evaluate their effects. 
%We present key results in the main text and additional results in the technical appendix. 

% As sensor failures are quite rare (only about $0.05\%$ in a year), we observed that our randomly sampled data consisted of minimal failures, if any. As a result, we first present results without any failures and then manually inject failures (randomly) while varying other environmental conditions to test the robustness of our approach.  
%Also, referring back to \cref{fig:SMDP}, the scene (location and weather) simultaneously changes anytime in the interval of $15$ to $45$ seconds ($t_{l}$ in the figure). 
%\input{graphs/travel_time_boxplot}

%\input{graphs/infraction_boxplot}

%\input{graphs/new-travel}

%\input{graphs/new-infractions}

% We also set the blur, occlusion, and novelty detector once every decision epoch and the collision likelihood estimator every inference cycle.

\textbf{\ac{av} operation with no sensor failure}: We begin by evaluating the performance of the controller configurations across all the tracks \textit{without any sensor failures}. As our objective is to maximize the system's performance while maintaining safety, we show results in terms of time taken to complete the track and the number of infractions (e.g., collisions). We begin by investigating the travel times taken by the controllers in \cref{fig:travel-safe} (top row).
% We define travel time as the time the vehicle takes to complete driving around a given track. The travel times of the different controller configurations around the three tracks are provided in \cref{fig:travel-time}. 

We first observe that the \ac{lbc} controller (the performant controller in isolation) has the best (i.e., lowest) travel times for all tracks aside from track 1. However, this performance comes at the cost of a highly unreliable safety performance (i.e., high variance in infraction scores, described below). The \ac{lbc} also fails to complete track 1 even once in the $30$ runs due to catastrophic failures. We observe that the \ac{ap} controller can be erratic; while it (in isolation) often prioritizes safety resulting in track completion (with long travel times), it also leads to several failures, e.g., in track 2. We also observe that \textbf{the proposed approach ($DS$) results in the fewest infractions while achieving competitive travel times}. Note that longer travel times shown by $DS$ are also a result of prioritizing safety, which we describe next.

In order to evaluate safety, we compute an \textit{infraction score} as $0.5 \cdot RC + 0.25 \cdot col_{v} + 0.25 \cdot col_{o}$, where $RC$ is the percentage of the route that was completed by the vehicle, and $col_{v}$ and $col_{o}$ denote the presence of collisions with other vehicles and other objects (e.g., a wall) respectively. We design the score such that a \textbf{higher score is better}, i.e., if any infraction is observed, the resulting parameter in the score is set to 0; otherwise, it is set to 1. We observe that the traditional simplex configuration $SA$ is the safest among all the simplex configurations. We also observe that the $DS$ configuration shows comparable performance with a median infraction score of $1.0$ across all tracks. Analyzing the results on performance and safety, \textbf{we observe that the proposed $DS$ outperforms other approaches in terms of balancing safety and performance}. For exmaple, in track 3, while \ac{lbc} shows lower travel times and only fails to complete the track once, it also has a low median infraction score of $0.75$, which indicates a significant number of collisions with the other vehicles or objects.
Finally, we also observe that $DS$ performs much fewer switches than $GS$, indicating our method can approach the nearly optimal time to switch to achieve performance objective without sacrificing safety.

\input{graphs/permanent_failure}
\input{graphs/sensitivity_graph}
\textbf{\ac{av} operation with permanent sensor failure}: To explicitly evaluate how the controllers perform under sensor failures, we randomly inject faults during our evaluation. Specifically, we simulate a center camera occlusion for the \ac{av} at random times and persist the failure once it occurs. We show the results in \cref{fig:failure}. We observe that such permanent camera occlusion severely affects the \ac{lbc} controller, causing it to collide in all cases. On the other hand, the \ac{ap} controller is unaffected by the occlusion. The proposed approach ($DS$) significantly outperforms $SA$ and \ac{ap} in travel time across all tracks but also achieves a median infraction score of $1$ with low variance, thereby considerably improving the other simplex configurations. Note that $DS$ has to perform slightly more reverse switches than $GS$ to achieve performance objective without sacrificing infraction score on Track 2 and Track 4.

% track 4 creates a good condition to observe the negative impact caused by the sensor failures on the all configurations independent of the other negative exogenous factors in the environment. 

% We simulate a center camera occlusion for the \ac{av} operating in track 4 (see \cref{fig:regions}),  We introduced the occlusion at random time into the simulation, which persists throughout the vehicle's operation. Our first observation is that such permanent camera occlusion severely affects the \ac{lbc} controller, causing it to collide all the $30$ times, resulting in a low median safety score of $0.44$. The \ac{ap} controller is unaffected by the occlusion. So, it has a perfect safety score of $1.0$ with a median travel time of $188$ seconds. We also observed that the simplex configurations frequently switched to the \ac{ap} controller to overcome the problem with the \ac{lbc} controller. Among these, the $DS$ has shorter median travel times of $109.15$ seconds, significantly improving on the other simplex configurations. This reduction is because the occlusion only affects the \ac{lbc} controller's performance in some scenes except the scene in which the \ac{lbc} controller failed. However, the other simplex configurations operate with the \ac{ap} controller for most of the track or frequently change between controllers causing frequent occurrences of switching routine.  

\input{graphs/intermittent_failure}
\textbf{\ac{av} operation with intermittent sensor failure}: As highlighted in the \cref{sec:evaluation} in the main text, we also simulate intermittent failures from which an autonomous CPS can recover. For example, for an AV, an intermittent failure can be caused by high precipitation or strong light affecting the camera (e.g., sunlight). In our experiments, we use an exponential growth function to simulate the likelihood of occlusion conditioned on weather and location, i.e., the sunnier or heavier the precipitation is, it is more likely to cause a temporary occlusion. We also ensure that such failures are dependent on the structural features of the state, e.g., sunny conditions or precipitation is unlikely to cause occlusion if the vehicle is operating in a tunnel. 
% \ava{you discuss how to model temporary occlusion arrival times, but how long do they persist?}

We present the results with intermittent sensor failure in \cref{fig:interfailure}. The intermittent failures do not affect the controllers as much as the permanent failures (\cref{fig:failure}). We observe that the $GS$ controller fails to complete the tracks significantly more times than the other controllers. All the other four controllers have similar performance as they have in \cref{fig:travel-safe} with regard to the ability to complete the tracks, and the $DS$ controller shows a lower median travel time while performing equally in terms of safety than $SA$. Though $GS$ offers much lower median travel time than $DS$ on Track2 and Track 4, it achieves so by significantly sacrificing the safety and ability to complete the tracks. 

% We can further observe that $DS$ has to perform slightly more reverse switches than $GS$ on all 4 tracks, indicating $DS$ can increase the number of MCTS iterations to mitigate such phenomenon if necessary. Nevertheless, $DS$ outperforms $SA$ and $GS$ in travel time and infraction scores, respectively.

\textbf{Sensitivity Analysis}:
Recall that in \cref{eqn:fwd} and \cref{eqn:rev}, $\alpha_1$, $\alpha_2$, and $\alpha_3$ denote the weights to trade off the objectives of performance, safety, and avoiding frequent switching, respectively. We perform sensitivity analysis to analyze the effects of these weights by keeping $\alpha_2$ fixed at 1 and varying $\alpha_1$ and $\alpha_3$. In \cref{fig:sensitivity}, we show how the performance of the proposed $DS$ controller without sensor failures is affected by the weights. We observe that increasing the value of $\alpha_3$ from $0$ to $1$ gradually reduces the number of switches. We also observe that small values of $\alpha_1$ generally lead to a lower performance score, which indicates longer travel times. Furthermore, we find that setting $\alpha_1$ to $1$ and $\alpha_3$ to $0$ results in the best performance score but comes with the sacrifice of infraction score, which implies the occurrence of failures or collisions. Finally, constraining the performance score and the number of switches by setting $\alpha_1$ to 0 and $\alpha_3$ to 1 (thereby penalizing switches heavily) sacrifices performance significantly, emphasizing the need of performing principled switching to balance safety and performance.
\input{tables/computation_time}

\textbf{Computation Time}:
\cref{tab:computation} shows the average computation times, travel time, infraction score, and the average number of switches taken by the decision logic of our proposed $DS$ configurations. Note that the average computation time highly correlates to the computational capacity of the hardware. Recall that the forward switching of the configuration is performed based on inference by a model trained using historical data. As a result, it is extremely fast, as required in practice for ensuring safety. The $DS$ uses the MCTS-based planner for reverse switching, which requires an average time proportional to the increased number of MCTS iterations. We observe that $500$ MCST iterations give the best average travel time without harming the system's safety. We also observe that the planner with $100$ MCTS iterations takes the least time to make decision while sacrificing the infraction score and increasing the number of switches. However, it can still offer a competitive average travel time compared to the planner with $1000$ and $2000$ MCTS iterations, enabling the decision-maker to put constraints on
the computational time and latency that can be afforded to make a
decision depending on domain-specific requirements.

%% file: graphs/travel_safe_boxplot.tex
\usepgfplotslibrary{statistics}
\pgfplotsset{
/pgfplots/custom legend/.style={
legend image code/.code={
\draw [only marks,mark=square]
plot coordinates { 
(0.3cm,0cm)
};
}, },
}

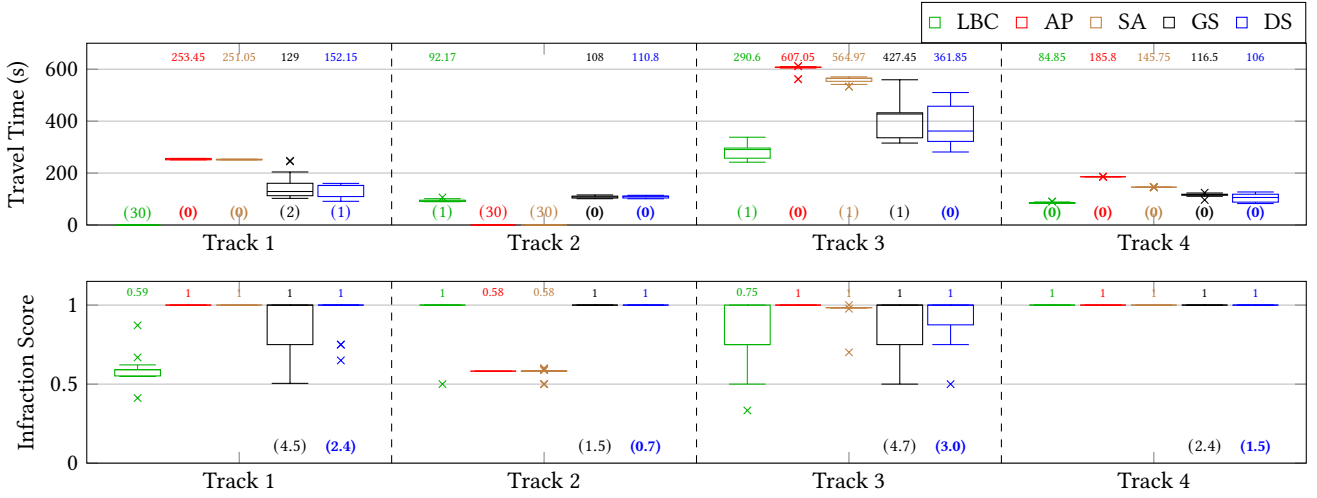
\begin{figure*}
\begin{tikzpicture}
\begin{groupplot}[group style = {group size = 1 by 2, horizontal sep = 2cm, vertical sep = 0.75cm }]

\nextgroupplot[
        height=4cm, width=\textwidth,
      %bugsResolvedStyle/.style={},
      %xlabel={Track 1},
      ylabel={Travel Time (s)},
     ymin=0,
     xmin=0,
     ymax = 700,
     xmax = 6,
     xtick={0.75,2.25,3.75,5.25},
     xticklabels={{Track 1},{Track 2},{Track 3},{Track 4}},
     ymajorgrids,
     boxplot/draw direction=y,
     cycle list={{red},{blue}},
        %enlarge y limits=.2,
    %/pgfplots/boxplot/box extend=0.3,
      custom legend,
        legend style={at={(1,1)},anchor=south east,legend columns=6,
            column sep=0.5em}
]

% \pgfplotsset{
%     boxplot/draw/median/.code={
%         \draw [/pgfplots/boxplot/every median/.try]
%             (boxplot box cs:\pgfplotsboxplotvalue{median},0)
%             node[anchor=north west, font=\tiny] {\pgfmathprintnumber{\pgfplotsboxplotvalue{median}}}
%             (boxplot box cs:\pgfplotsboxplotvalue{median},1);
%     },
% }

\addlegendentry{LBC}
\addlegendentry{AP}
\addlegendentry{SA}
\addlegendentry{GS}
\addlegendentry{DS}

\node at (axis cs:0.25,-20) [anchor=south][black!30!green] {\scriptsize $(30)$};
\node at (axis cs:0.5,-15) [anchor=south][red] {\scriptsize {$\textbf{(0)}$}};
\node at (axis cs:0.75,-15) [anchor=south][brown] {\scriptsize {$\textbf{(0)}$}};
\node at (axis cs:1,-15) [anchor=south][black] {\scriptsize $(2)$};
\node at (axis cs:1.25,-15) [anchor=south][blue] {\scriptsize $(1)$};

%\node at (axis cs:1,282) [anchor=north] [black!30!green] {\tiny 273};
\node at (axis cs:0.5,600) [anchor=south] [red] {\tiny 253.45};
\node at (axis cs:0.75,600) [anchor=south] [brown] {\tiny 251.05};
\node at (axis cs:1,600) [anchor=south] [black] {\tiny 129};
\node at (axis cs:1.25,600) [anchor=south] [blue] {\tiny 152.15};    

\node at (axis cs:1.75,600) [anchor=south] [black!30!green] {\tiny 92.17};
\node at (axis cs:2.5,600) [anchor=south] [black] {\tiny 108};
\node at (axis cs:2.75,600) [anchor=south] [blue] {\tiny 110.8};

\node at (axis cs:3.25,600) [anchor=south] [black!30!green] {\tiny 290.6};
\node at (axis cs:3.5,600) [anchor=south] [red] {\tiny 607.05};
\node at (axis cs:3.75,600) [anchor=south] [brown] {\tiny 564.97};
\node at (axis cs:4,600) [anchor=south] [black] {\tiny 427.45};  
\node at (axis cs:4.25,600) [anchor=south] [blue] {\tiny 361.85};  

\node at (axis cs:4.75,600) [anchor=south] [black!30!green] {\tiny 84.85};
\node at (axis cs:5,600) [anchor=south] [red] {\tiny 185.8};
\node at (axis cs:5.25,600) [anchor=south] [brown] {\tiny 145.75};
\node at (axis cs:5.5,600) [anchor=south] [black] {\tiny 116.5};  
\node at (axis cs:5.75,600) [anchor=south] [blue] {\tiny 106};  
    
\addplot+ [black!30!green][boxplot={box extend=0.225, draw position=0.25},mark=x, mark=x] table [y index=0] {data/lbc/Track1_time.txt};

\addplot+ [red][boxplot={box extend=0.225, draw position=0.5},mark=x] table [y index=0] {data/ap/Track4_time.txt};
  % Ignore the miscalculations here
\addplot+ [brown][boxplot={box extend=0.225, draw position=0.75},mark=x] table [y index=0] {data/sa/Track4_time.txt};

\addplot+ [black][boxplot={box extend=0.225, draw position=1},mark=x] table [y index=0] {data/sag/Track1_time.txt};

\addplot+ [blue][boxplot={box extend=0.225, draw position=1.25},mark=x] table [y index=0] {data/dss/Track1_time.txt};
%%%%%%%%%%%%%%%%%%%%%%%%%%%%%%%%%%%%%%%%%%%%%%%%%%%%%%%%%%
\node at (axis cs:1.75,-15) [anchor=south][black!30!green] {\scriptsize $(1)$};
\node at (axis cs:2,-15) [anchor=south][red] {\scriptsize $(30)$};
\node at (axis cs:2.25,-15) [anchor=south][brown] {\scriptsize $(30)$};
\node at (axis cs:2.5,-15) [anchor=south][black] {\scriptsize {$\textbf{(0)}$}};
\node at (axis cs:2.75,-15) [anchor=south][blue] {\scriptsize {$\textbf{(0)}$}};

\addplot+ [black!30!green][boxplot={box extend=0.225, draw position=1.75}, mark=x] table [y index=0] {data/lbc/Track2_time.txt};

\addplot+ [red][boxplot={box extend=0.225, draw position=2},mark=x] table [y index=0] {data/ap/Track1_time.txt};

\addplot+ [brown][boxplot={box extend=0.225, draw position=2.25},mark=x] table [y index=0] {data/sa/Track1_time.txt};

\addplot+ [black][boxplot={box extend=0.225, draw position=2.5},mark=x] table [y index=0] {data/sag/Track2_time.txt};
\addplot+ [blue][boxplot={box extend=0.225, draw position=2.75},mark=x] table [y index=0] {data/dss/Track2_time.txt};    

%%%%%%%%%%%%%%%%%%%%%%%%%%%%%%%%%%%%%%%%%%%%%%%%%%%%%%%%%%%%%%%%%
\node at (axis cs:3.25,-15) [anchor=south][black!30!green] {\scriptsize $(1)$};
\node at (axis cs:3.5,-15) [anchor=south][red] {\scriptsize {$\textbf{(0)}$}};
\node at (axis cs:3.75,-15) [anchor=south][brown] {\scriptsize $(1)$};
\node at (axis cs:4.0,-15) [anchor=south][black] {\scriptsize $(1)$};
\node at (axis cs:4.25,-15) [anchor=south][blue] {\scriptsize {$\textbf{(0)}$}};

\addplot+ [black!30!green][boxplot={box extend=0.225, draw position=3.25},mark=x] table [y index=0] {data/lbc/Track3_time.txt};

\addplot+ [red][boxplot={box extend=0.225, draw position=3.5},mark=x] table [y index=0] {data/ap/Track3_time.txt};
  % Ignore the miscalculations here
\addplot+ [brown][boxplot={box extend=0.225, draw position=3.75},mark=x] table [y index=0] {data/sa/Track3_time.txt};

\addplot+ [black][boxplot={box extend=0.225, draw position=4.0},mark=x] table [y index=0] {data/sag/Track3_time.txt};

\addplot+ [blue][boxplot={box extend=0.225, draw position=4.25},mark=x] table [y index=0] {data/dss/Track3_time.txt};
%%%%%%%%%%%%%%%%%%%%%%%%%%%%%%%%%%%%%%%%%%%%%%%%%%%%%%%%%%%%%%%%
\node at (axis cs:4.75,-15) [anchor=south][black!30!green] {\scriptsize $\textbf{(0)}$};
\node at (axis cs:5,-15) [anchor=south][red] {\scriptsize {$\textbf{(0)}$}};
\node at (axis cs:5.25,-15) [anchor=south][brown] {\scriptsize {$\textbf{(0)}$}};
\node at (axis cs:5.5,-15) [anchor=south][black] {\scriptsize {$\textbf{(0)}$}};
\node at (axis cs:5.75,-15) [anchor=south][blue] {\scriptsize {$\textbf{(0)}$}};

\addplot+ [black!30!green][boxplot={box extend=0.225, draw position=4.75},mark=x] table [y index=0] {data/lbc/Track4_time.txt};
\addplot+ [boxplot={box extend=0.225, draw position=5},mark=x] table [y index=0] {data/ap/Track2_time.txt};
  % Ignore the miscalculations here
\addplot+ [brown][boxplot={box extend=0.225, draw position=5.25},mark=x] table [y index=0] {data/sa/Track2_time.txt};
\addplot+ [black][boxplot={box extend=0.225, draw position=5.5},mark=x] table [y index=0] {data/sag/Track4_time.txt};

\addplot+ [blue][boxplot={box extend=0.225, draw position=5.75},mark=x] table [y index=0] {data/dss/Track4_time.txt};

\draw [dashed] (1.5,0) -- (1.5,700);
\draw [dashed] (3,0) -- (3,700);
\draw [dashed] (4.5,0) -- (4.5,700);
]

\nextgroupplot[
        height=4cm, width=\textwidth,
      %bugsResolvedStyle/.style={},
      %xlabel={Track 1},
      ylabel={Infraction Score},
     ymin=0,
     xmin=0,
     ymax = 1.15,
     xmax = 6,
     xtick={0.75,2.25,3.75,5.25},
     xticklabels={{Track 1},{Track 2},{Track 3},{Track 4}},
     ymajorgrids,
     boxplot/draw direction=y,
     cycle list={{red},{blue}},
        %enlarge y limits=.2,
    %/pgfplots/boxplot/box extend=0.3,
      custom legend,
        legend style={at={(1,1)},anchor=south east,legend columns=6,
            column sep=0.5em}
]

%\pgfplotsset{
%     boxplot/draw/median/.code={
%         \draw [/pgfplots/boxplot/every median/.try]
%             (boxplot box cs:\pgfplotsboxplotvalue{median},0)
%             node[anchor=north west, font=\tiny] {\pgfmathprintnumber{\pgfplotsboxplotvalue{median}}}
%             (boxplot box cs:\pgfplotsboxplotvalue{median},1);
%     },
% }

\node at (axis cs:1,0) [anchor=south][black] {\scriptsize $(4.5)$};
\node at (axis cs:1.25,0) [anchor=south][blue] {\scriptsize $\textbf{(2.4)}$};

\node at (axis cs:2.5,0) [anchor=south][black] {\scriptsize $(1.5)$};
\node at (axis cs:2.75,0) [anchor=south][blue] {\scriptsize $\textbf{(0.7)}$};

\node at (axis cs:4,0) [anchor=south][black] {\scriptsize $(4.7)$};
\node at (axis cs:4.25,0) [anchor=south][blue] {\scriptsize $\textbf{(3.0)}$};

\node at (axis cs:5.5,0) [anchor=south][black] {\scriptsize $(2.4)$};
\node at (axis cs:5.75,0) [anchor=south][blue] {\scriptsize $\textbf{(1.5)}$};

\node at (axis cs:0.25,1) [anchor=south] [black!30!green] {\tiny 0.59};
\node at (axis cs:0.5,1) [anchor=south] [red] {\tiny 1};
\node at (axis cs:0.75,1) [anchor=south] [brown] {\tiny 1};
\node at (axis cs:1,1) [anchor=south] [black] {\tiny 1};
\node at (axis cs:1.25,1) [anchor=south] [blue] {\tiny 1};    

\node at (axis cs:1.75,1) [anchor=south] [black!30!green] {\tiny 1};
\node at (axis cs:2,1) [anchor=south] [red] {\tiny 0.58};
\node at (axis cs:2.25,1) [anchor=south] [brown] {\tiny 0.58};
\node at (axis cs:2.5,1) [anchor=south] [black] {\tiny 1};
\node at (axis cs:2.75,1) [anchor=south] [blue] {\tiny 1};

\node at (axis cs:3.25,1) [anchor=south] [black!30!green] {\tiny 0.75};
\node at (axis cs:3.5,1) [anchor=south] [red] {\tiny 1};
\node at (axis cs:3.75,1) [anchor=south] [brown] {\tiny 1};
\node at (axis cs:4,1) [anchor=south] [black] {\tiny 1};  
\node at (axis cs:4.25,1) [anchor=south] [blue] {\tiny 1};  

\node at (axis cs:4.75,1) [anchor=south] [black!30!green] {\tiny 1};
\node at (axis cs:5,1) [anchor=south] [red] {\tiny 1};
\node at (axis cs:5.25,1) [anchor=south] [brown] {\tiny 1};
\node at (axis cs:5.5,1) [anchor=south] [black] {\tiny 1};  
\node at (axis cs:5.75,1) [anchor=south] [blue] {\tiny 1};

\addplot+ [black!30!green][boxplot={box extend=0.225, draw position=0.25},mark=x, mark=x] table [y index=0] {data/lbc/Track1_safe.txt};

\addplot+ [red][boxplot={box extend=0.225, draw position=0.5},mark=x] table [y index=0] {data/ap/Track4_safe.txt};
  % Ignore the miscalculations here
\addplot+ [brown][boxplot={box extend=0.225, draw position=0.75},mark=x] table [y index=0] {data/sa/Track4_safe.txt};

\addplot+ [black][boxplot={box extend=0.225, draw position=1},mark=x] table [y index=0] {data/sag/Track1_safe.txt};

\addplot+ [blue][boxplot={box extend=0.225, draw position=1.25},mark=x] table [y index=0] {data/dss/Track1_safe.txt};
%%%%%%%%%%%%%%%%%%%%%%%%%%%%%%%%%%%%%%%%%%%%%%%%%%%%%%%%%%

\addplot+ [black!30!green][boxplot={box extend=0.225, draw position=1.75}, mark=x] table [y index=0] {data/lbc/Track2_safe.txt};

\addplot+ [red][boxplot={box extend=0.225, draw position=2},mark=x] table [y index=0] {data/ap/Track1_safe.txt};

\addplot+ [brown][boxplot={box extend=0.225, draw position=2.25},mark=x] table [y index=0] {data/sa/Track1_safe.txt};

\addplot+ [black][boxplot={box extend=0.225, draw position=2.5},mark=x] table [y index=0] {data/sag/Track2_safe.txt};
\addplot+ [blue][boxplot={box extend=0.225, draw position=2.75},mark=x] table [y index=0] {data/dss/Track2_safe.txt};    

%%%%%%%%%%%%%%%%%%%%%%%%%%%%%%%%%%%%%%%%%%%%%%%%%%%%%%%%%%%%%%%%%

\addplot+ [black!30!green][boxplot={box extend=0.225, draw position=3.25},mark=x] table [y index=0] {data/lbc/Track3_safe.txt};

\addplot+ [red][boxplot={box extend=0.225, draw position=3.5},mark=x] table [y index=0] {data/ap/Track3_safe.txt};
  % Ignore the miscalculations here
\addplot+ [brown][boxplot={box extend=0.225, draw position=3.75},mark=x] table [y index=0] {data/sa/Track3_safe.txt};

\addplot+ [black][boxplot={box extend=0.225, draw position=4.0},mark=x] table [y index=0] {data/sag/Track3_safe.txt};

\addplot+ [blue][boxplot={box extend=0.225, draw position=4.25},mark=x] table [y index=0] {data/dss/Track3_safe.txt};
%%%%%%%%%%%%%%%%%%%%%%%%%%%%%%%%%%%%%%%%%%%%%%%%%%%%%%%%%%%%%%%%

\addplot+ [black!30!green][boxplot={box extend=0.225, draw position=4.75},mark=x] table [y index=0] {data/lbc/Track4_safe.txt};
\addplot+ [red][boxplot={box extend=0.225, draw position=5},mark=x] table [y index=0] {data/ap/Track2_safe.txt};
  % Ignore the miscalculations here
\addplot+ [brown][boxplot={box extend=0.225, draw position=5.25},mark=x] table [y index=0] {data/sa/Track2_safe.txt};
\addplot+ [black][boxplot={box extend=0.225, draw position=5.5},mark=x] table [y index=0] {data/sag/Track4_safe.txt};

\addplot+ [blue][boxplot={box extend=0.225, draw position=5.75},mark=x] table [y index=0] {data/dss/Track4_safe.txt};

\draw [dashed] (1.5,0) -- (1.5,1.2);
\draw [dashed] (3,0) -- (3,1.2);
\draw [dashed] (4.5,0) -- (4.5,1.2);
]
\end{groupplot}
\end{tikzpicture}
\caption{\textbf{Top:} Travel times of the different controller configurations across the $4$ tracks \textbf{(lower is better)}. We show the number of times a controller failed to complete a track in parenthesis below each box, highlighting the least failures in bold. We also show the median of the distributions at the top of the graph. We observe $DS$ is the only configuration that provides competitive performance without sacrificing safety. \textbf{(Bottom):} We show the infraction score \textbf{(higher is better)} for all controllers across all tracks. We observe that $DS$ consistently achieves a median score of 1, the highest possible safety score. We also show the mean of reverse switches performed by $GS$ and $DS$ across all tracks below each box. We observe that $DS$ performs significantly fewer switches than $GS$.}
\label{fig:travel-safe}
\end{figure*}

%the method works fairly well 

%% file: graphs/permanent_failure.tex
\usepgfplotslibrary{statistics}
\pgfplotsset{
/pgfplots/custom legend/.style={
legend image code/.code={
\draw [only marks,mark=square]
plot coordinates { 
(0.3cm,0cm)
};
}, },
}

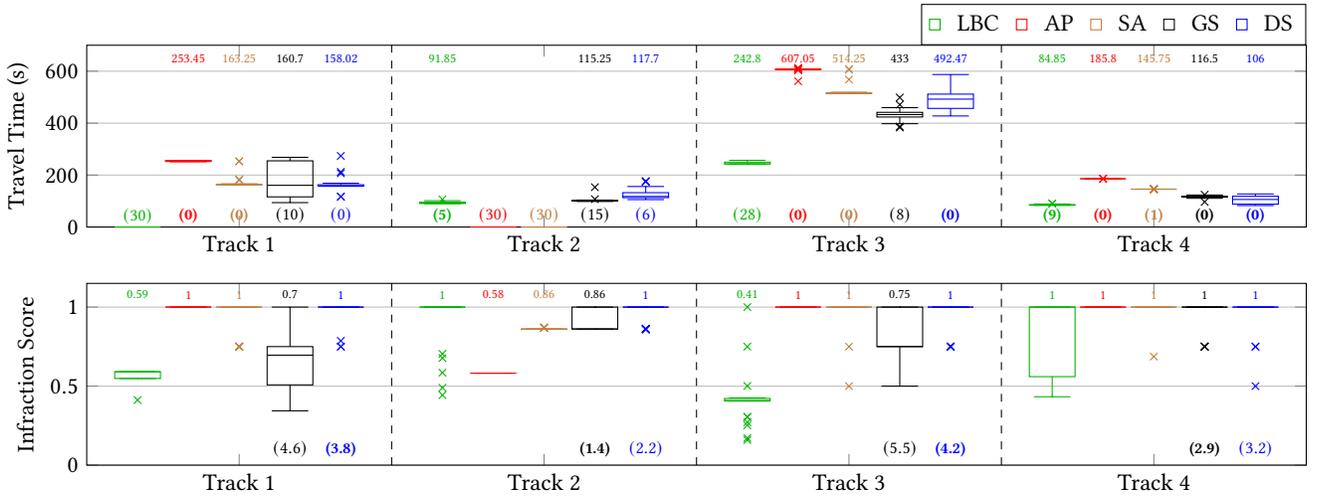
\begin{figure*}
\begin{tikzpicture}
\begin{groupplot}[group style = {group size = 1 by 2, horizontal sep = 2cm, vertical sep = 0.75cm }]

\nextgroupplot[
        height=4cm, width=\textwidth,
      %bugsResolvedStyle/.style={},
      %xlabel={Track 1},
      ylabel={Travel Time (s)},
     ymin=0,
     xmin=0,
     ymax = 700,
     xmax = 6,
     xtick={0.75,2.25,3.75,5.25},
     xticklabels={{Track 1},{Track 2},{Track 3},{Track 4}},
     ymajorgrids,
     boxplot/draw direction=y,
     cycle list={{red},{blue}},
        %enlarge y limits=.2,
    %/pgfplots/boxplot/box extend=0.3,
      custom legend,
        legend style={at={(1,1)},anchor=south east,legend columns=6,
            column sep=0.5em}
]

% \pgfplotsset{
%     boxplot/draw/median/.code={
%         \draw [/pgfplots/boxplot/every %median/.try]
%             (boxplot box cs:\pgfplotsboxplotvalue{median},0)
%             node[anchor=north west, font=\tiny] {\pgfmathprintnumber{\pgfplotsboxplotvalue{median}}}
%             (boxplot box cs:\pgfplotsboxplotvalue{median},1);
%     },
% }

\addlegendentry{LBC}
\addlegendentry{AP}
\addlegendentry{SA}
\addlegendentry{GS}
\addlegendentry{DS}

\node at (axis cs:0.25,-20) [anchor=south][black!30!green] {\scriptsize $(30)$};
\node at (axis cs:0.5,-15) [anchor=south][red] {\scriptsize {$\textbf{(0)}$}};
\node at (axis cs:0.75,-15) [anchor=south][brown] {\scriptsize {$\textbf{(0)}$}};
\node at (axis cs:1,-15) [anchor=south][black] {\scriptsize $(10)$};
\node at (axis cs:1.25,-15) [anchor=south][blue] {\scriptsize $(0)$};

%\node at (axis cs:1,282) [anchor=north] [black!30!green] {\tiny 273};
\node at (axis cs:0.5,600) [anchor=south] [red] {\tiny 253.45};
\node at (axis cs:0.75,600) [anchor=south] [brown] {\tiny 163.25};
\node at (axis cs:1,600) [anchor=south] [black] {\tiny 160.7};
\node at (axis cs:1.25,600) [anchor=south] [blue] {\tiny 158.02};    

\node at (axis cs:1.75,600) [anchor=south] [black!30!green] {\tiny 91.85};
\node at (axis cs:2.5,600) [anchor=south] [black] {\tiny 115.25};
\node at (axis cs:2.75,600) [anchor=south] [blue] {\tiny 117.7};

\node at (axis cs:3.25,600) [anchor=south] [black!30!green] {\tiny 242.8};
\node at (axis cs:3.5,600) [anchor=south] [red] {\tiny 607.05};
\node at (axis cs:3.75,600) [anchor=south] [brown] {\tiny 514.25};
\node at (axis cs:4,600) [anchor=south] [black] {\tiny 433};  
\node at (axis cs:4.25,600) [anchor=south] [blue] {\tiny 492.47};  

\node at (axis cs:4.75,600) [anchor=south] [black!30!green] {\tiny 84.85};
\node at (axis cs:5,600) [anchor=south] [red] {\tiny 185.8};
\node at (axis cs:5.25,600) [anchor=south] [brown] {\tiny 145.75};
\node at (axis cs:5.5,600) [anchor=south] [black] {\tiny 116.5};  
\node at (axis cs:5.75,600) [anchor=south] [blue] {\tiny 106};  
    
\addplot+ [black!30!green][boxplot={box extend=0.225, draw position=0.25},mark=x, mark=x] table [y index=0] {failure_data/lbc/Track1_time.txt};

\addplot+ [red][boxplot={box extend=0.225, draw position=0.5},mark=x] table [y index=0] {failure_data/ap/Track1_time.txt};
  % Ignore the miscalculations here
\addplot+ [brown][boxplot={box extend=0.225, draw position=0.75},mark=x] table [y index=0] {failure_data/sa/Track1_time.txt};

\addplot+ [black][boxplot={box extend=0.225, draw position=1},mark=x] table [y index=0] {failure_data/sag/Track1_time.txt};

\addplot+ [blue][boxplot={box extend=0.225, draw position=1.25},mark=x] table [y index=0] {failure_data/dss/Track1_time.txt};
%%%%%%%%%%%%%%%%%%%%%%%%%%%%%%%%%%%%%%%%%%%%%%%%%%%%%%%%%%
\node at (axis cs:1.75,-15) [anchor=south][black!30!green] {\scriptsize $\textbf{(5)}$};
\node at (axis cs:2,-15) [anchor=south][red] {\scriptsize $(30)$};
\node at (axis cs:2.25,-15) [anchor=south][brown] {\scriptsize $(30)$};
\node at (axis cs:2.5,-15) [anchor=south][black] {\scriptsize {$(15)$}};
\node at (axis cs:2.75,-15) [anchor=south][blue] {\scriptsize {$(6)$}};

\addplot+ [black!30!green][boxplot={box extend=0.225, draw position=1.75}, mark=x] table [y index=0] {failure_data/lbc/Track2_time.txt};

\addplot+ [red][boxplot={box extend=0.225, draw position=2},mark=x] table [y index=0] {failure_data/ap/Track2_time.txt};

\addplot+ [brown][boxplot={box extend=0.225, draw position=2.25},mark=x] table [y index=0] {failure_data/sa/Track2_time.txt};

\addplot+ [black][boxplot={box extend=0.225, draw position=2.5},mark=x] table [y index=0] {failure_data/sag/Track2_time.txt};
\addplot+ [blue][boxplot={box extend=0.225, draw position=2.75},mark=x] table [y index=0] {failure_data/dss/Track2_time.txt};    

%%%%%%%%%%%%%%%%%%%%%%%%%%%%%%%%%%%%%%%%%%%%%%%%%%%%%%%%%%%%%%%%%
\node at (axis cs:3.25,-15) [anchor=south][black!30!green] {\scriptsize $(28)$};
\node at (axis cs:3.5,-15) [anchor=south][red] {\scriptsize {$\textbf{(0)}$}};
\node at (axis cs:3.75,-15) [anchor=south][brown] {\scriptsize $\textbf{(0)}$};
\node at (axis cs:4.0,-15) [anchor=south][black] {\scriptsize $(8)$};
\node at (axis cs:4.25,-15) [anchor=south][blue] {\scriptsize {$\textbf{(0)}$}};

\addplot+ [black!30!green][boxplot={box extend=0.225, draw position=3.25},mark=x] table [y index=0] {failure_data/lbc/Track3_time.txt};

\addplot+ [red][boxplot={box extend=0.225, draw position=3.5},mark=x] table [y index=0] {failure_data/ap/Track3_time.txt};
  % Ignore the miscalculations here
\addplot+ [brown][boxplot={box extend=0.225, draw position=3.75},mark=x] table [y index=0] {failure_data/sa/Track3_time.txt};

\addplot+ [black][boxplot={box extend=0.225, draw position=4.0},mark=x] table [y index=0] {failure_data/sag/Track3_time.txt};

\addplot+ [blue][boxplot={box extend=0.225, draw position=4.25},mark=x] table [y index=0] {failure_data/dss/Track3_time.txt};
%%%%%%%%%%%%%%%%%%%%%%%%%%%%%%%%%%%%%%%%%%%%%%%%%%%%%%%%%%%%%%%%
\node at (axis cs:4.75,-15) [anchor=south][black!30!green] {\scriptsize $\textbf{(9)}$};
\node at (axis cs:5,-15) [anchor=south][red] {\scriptsize {$\textbf{(0)}$}};
\node at (axis cs:5.25,-15) [anchor=south][brown] {\scriptsize {$\textbf{(1)}$}};
\node at (axis cs:5.5,-15) [anchor=south][black] {\scriptsize {$\textbf{(0)}$}};
\node at (axis cs:5.75,-15) [anchor=south][blue] {\scriptsize {$\textbf{(0)}$}};

\addplot+ [black!30!green][boxplot={box extend=0.225, draw position=4.75},mark=x] table [y index=0] {data/lbc/Track4_time.txt};
\addplot+ [boxplot={box extend=0.225, draw position=5},mark=x] table [y index=0] {data/ap/Track2_time.txt};
  % Ignore the miscalculations here
\addplot+ [brown][boxplot={box extend=0.225, draw position=5.25},mark=x] table [y index=0] {data/sa/Track2_time.txt};
\addplot+ [black][boxplot={box extend=0.225, draw position=5.5},mark=x] table [y index=0] {data/sag/Track4_time.txt};

\addplot+ [blue][boxplot={box extend=0.225, draw position=5.75},mark=x] table [y index=0] {data/dss/Track4_time.txt};

\draw [dashed] (1.5,0) -- (1.5,700);
\draw [dashed] (3,0) -- (3,700);
\draw [dashed] (4.5,0) -- (4.5,700);
]

\nextgroupplot[
        height=4cm, width=\textwidth,
      %bugsResolvedStyle/.style={},
      %xlabel={Track 1},
      ylabel={Infraction Score},
     ymin=0,
     xmin=0,
     ymax = 1.15,
     xmax = 6,
     xtick={0.75,2.25,3.75,5.25},
     xticklabels={{Track 1},{Track 2},{Track 3},{Track 4}},
     ymajorgrids,
     boxplot/draw direction=y,
     cycle list={{red},{blue}},
        %enlarge y limits=.2,
    %/pgfplots/boxplot/box extend=0.3,
      custom legend,
        legend style={at={(1,1)},anchor=south east,legend columns=6,
            column sep=0.5em}
]

%\pgfplotsset{
%     boxplot/draw/median/.code={
%         \draw [/pgfplots/boxplot/every median/.try]
%             (boxplot box cs:\pgfplotsboxplotvalue{median},0)
%             node[anchor=north west, font=\tiny] {\pgfmathprintnumber{\pgfplotsboxplotvalue{median}}}
%            (boxplot box cs:\pgfplotsboxplotvalue{median},1);
%     },
% }

\node at (axis cs:1,0) [anchor=south][black] {\scriptsize $(4.6)$};
\node at (axis cs:1.25,0) [anchor=south][blue] {\scriptsize $\textbf{(3.8)}$};

\node at (axis cs:2.5,0) [anchor=south][black] {\scriptsize $\textbf{(1.4)}$};
\node at (axis cs:2.75,0) [anchor=south][blue] {\scriptsize $(2.2)$};

\node at (axis cs:4,0) [anchor=south][black] {\scriptsize $(5.5)$};
\node at (axis cs:4.25,0) [anchor=south][blue] {\scriptsize $\textbf{(4.2)}$};

\node at (axis cs:5.5,0) [anchor=south][black] {\scriptsize $\textbf{(2.9)}$};
\node at (axis cs:5.75,0) [anchor=south][blue] {\scriptsize $(3.2)$};

\node at (axis cs:0.25,1) [anchor=south] [black!30!green] {\tiny 0.59};
\node at (axis cs:0.5,1) [anchor=south] [red] {\tiny 1};
\node at (axis cs:0.75,1) [anchor=south] [brown] {\tiny 1};
\node at (axis cs:1,1) [anchor=south] [black] {\tiny 0.7};
\node at (axis cs:1.25,1) [anchor=south] [blue] {\tiny 1};    

\node at (axis cs:1.75,1) [anchor=south] [black!30!green] {\tiny 1};
\node at (axis cs:2,1) [anchor=south] [red] {\tiny 0.58};
\node at (axis cs:2.25,1) [anchor=south] [brown] {\tiny 0.86};
\node at (axis cs:2.5,1) [anchor=south] [black] {\tiny 0.86};
\node at (axis cs:2.75,1) [anchor=south] [blue] {\tiny 1};

\node at (axis cs:3.25,1) [anchor=south] [black!30!green] {\tiny 0.41};
\node at (axis cs:3.5,1) [anchor=south] [red] {\tiny 1};
\node at (axis cs:3.75,1) [anchor=south] [brown] {\tiny 1};
\node at (axis cs:4,1) [anchor=south] [black] {\tiny 0.75};  
\node at (axis cs:4.25,1) [anchor=south] [blue] {\tiny 1};  

\node at (axis cs:4.75,1) [anchor=south] [black!30!green] {\tiny 1};
\node at (axis cs:5,1) [anchor=south] [red] {\tiny 1};
\node at (axis cs:5.25,1) [anchor=south] [brown] {\tiny 1};
\node at (axis cs:5.5,1) [anchor=south] [black] {\tiny 1};  
\node at (axis cs:5.75,1) [anchor=south] [blue] {\tiny 1};

\addplot+ [black!30!green][boxplot={box extend=0.225, draw position=0.25},mark=x, mark=x] table [y index=0] {failure_data/lbc/Track1_safe.txt};

\addplot+ [red][boxplot={box extend=0.225, draw position=0.5},mark=x] table [y index=0] {failure_data/ap/Track1_safe.txt};
  % Ignore the miscalculations here
\addplot+ [brown][boxplot={box extend=0.225, draw position=0.75},mark=x] table [y index=0] {failure_data/sa/Track1_safe.txt};

\addplot+ [black][boxplot={box extend=0.225, draw position=1},mark=x] table [y index=0] {failure_data/sag/Track1_safe.txt};

\addplot+ [blue][boxplot={box extend=0.225, draw position=1.25},mark=x] table [y index=0] {failure_data/dss/Track1_safe.txt};
%%%%%%%%%%%%%%%%%%%%%%%%%%%%%%%%%%%%%%%%%%%%%%%%%%%%%%%%%%

\addplot+ [black!30!green][boxplot={box extend=0.225, draw position=1.75}, mark=x] table [y index=0] {failure_data/lbc/Track2_safe.txt};

\addplot+ [red][boxplot={box extend=0.225, draw position=2},mark=x] table [y index=0] {failure_data/ap/Track2_safe.txt};

\addplot+ [brown][boxplot={box extend=0.225, draw position=2.25},mark=x] table [y index=0] {failure_data/sa/Track2_safe.txt};

\addplot+ [black][boxplot={box extend=0.225, draw position=2.5},mark=x] table [y index=0] {failure_data/sag/Track2_safe.txt};
\addplot+ [blue][boxplot={box extend=0.225, draw position=2.75},mark=x] table [y index=0] {failure_data/dss/Track2_safe.txt};    

%%%%%%%%%%%%%%%%%%%%%%%%%%%%%%%%%%%%%%%%%%%%%%%%%%%%%%%%%%%%%%%%%

\addplot+ [black!30!green][boxplot={box extend=0.225, draw position=3.25},mark=x] table [y index=0] {failure_data/lbc/Track3_safe.txt};

\addplot+ [red][boxplot={box extend=0.225, draw position=3.5},mark=x] table [y index=0] {failure_data/ap/Track3_safe.txt};
  % Ignore the miscalculations here
\addplot+ [brown][boxplot={box extend=0.225, draw position=3.75},mark=x] table [y index=0] {failure_data/sa/Track3_safe.txt};

\addplot+ [black][boxplot={box extend=0.225, draw position=4.0},mark=x] table [y index=0] {failure_data/sag/Track3_safe.txt};

\addplot+ [blue][boxplot={box extend=0.225, draw position=4.25},mark=x] table [y index=0] {failure_data/dss/Track3_safe.txt};
%%%%%%%%%%%%%%%%%%%%%%%%%%%%%%%%%%%%%%%%%%%%%%%%%%%%%%%%%%%%%%%%

\addplot+ [black!30!green][boxplot={box extend=0.225, draw position=4.75},mark=x] table [y index=0] {failure_data/lbc/Track4_safe.txt};
\addplot+ [red][boxplot={box extend=0.225, draw position=5},mark=x] table [y index=0] {failure_data/ap/Track4_safe.txt};
  % Ignore the miscalculations here
\addplot+ [brown][boxplot={box extend=0.225, draw position=5.25},mark=x] table [y index=0] {failure_data/sa/Track4_safe.txt};
\addplot+ [black][boxplot={box extend=0.225, draw position=5.5},mark=x] table [y index=0] {failure_data/sag/Track4_safe.txt};

\addplot+ [blue][boxplot={box extend=0.225, draw position=5.75},mark=x] table [y index=0] {failure_data/dss/Track4_safe.txt};

\draw [dashed] (1.5,0) -- (1.5,1.2);
\draw [dashed] (3,0) -- (3,1.2);
\draw [dashed] (4.5,0) -- (4.5,1.2);
]
\end{groupplot}
\end{tikzpicture}
\caption{\textbf{Top}: Travel times for all controllers with permanent camera occlusion on all 4 tracks \textbf{(lower is better)}. We show the number of times a controller failed to complete a track in parenthesis below each box, highlighting the least failures in bold. We also show the median of the distributions. \textbf{Bottom}: The infraction score (higher is better) for all controllers with permanent camera occlusion on all 4 tracks. We observe that DS consistently achieves a median score of 1, the highest possible safety score. We also show the mean of reverse switches performed by $GS$ and $DS$ across all tracks below each box.}
\label{fig:failure}
\end{figure*}

%the method works fairly well 

%% file: graphs/sensitivity_graph.tex
% Spiderweb Diagram
%
% Author: Dominik Renzel
% Date; 2009-11-11

\usetikzlibrary{shapes}

%\begin{document}

% replaced commas with periods
% added colour-column

\newcommand{\D}{3} % number of dimensions (config option)
\newcommand{\B}{10} % number of scale units (config option)

\newdimen\R % maximal diagram radius (config option)
\R=3.5cm 
\newdimen\L % radius to put dimension labels (config option)
\L=4cm

\newcommand{\A}{360/\D} % calculated angle between dimension axes  

\begin{figure}[!htb]
\centering

\begin{tikzpicture}[scale=0.75]
\path (0:0cm) coordinate (O); % define coordinate for origin

% draw the spiderweb
  \foreach \X in {1,...,\D}{
    \draw ({-(\X)*\A+90}:0) -- ({-(\X)*\A+90}:\R);
  }

  \foreach \Y in {0,...,\B}{
    \foreach \X in {1,...,\D}{
      \path ({-(\X)*\A+90}:\Y*\R/\B) coordinate (D\X-\Y);
      \fill (D\X-\Y) circle (1pt);
    }
    \draw [opacity=0.3] (90:\Y*\R/\B) \foreach \X in {1,...,\D}{
        -- ({-(\X)*\A+90}:\Y*\R/\B)
    } -- cycle;
  }

  % define labels for each dimension axis (names config option)
  \path (-1*\A+90:\L) node (L1) {\small Performance Score};
  \path (-2*\A+90:\L) node (L2) {\small Infraction Score};
  \path (-3*\A+90:\L) node (L3) {\small Switch Number};

% common scale factor for the values
\newcommand\datascale{3.1}

\csvreader[separator=semicolon]
    {data.csv}
    {parameter=\v,performance=\port,safety=\comp,
     switch=\perf,color=\clr}
    {
        \draw[\clr,line width=1.5pt,opacity=0.1]
            (-1*\A+90:\port/\datascale) -- 
            (-2*\A+90:\comp/\datascale) -- 
            (-3*\A+90:\perf/\datascale) -- cycle;
            %(\A*4:\use/\datascale) -- 
            %(\A*5:\secur/\datascale) -- 
            %(\A*6:\rel/\datascale) -- 
            %(\A*7:\func/\datascale) -- cycle;

       % add legend
       %\draw [line width=1.5pt,\clr,opacity=1] (\R*0.5,\R-12pt*\the\numexpr2+\thecsvrow\relax) -- ++(1,0) node[right,opacity=1,black] {\v};
    }
\csvreader[separator=semicolon]
    {data_2.csv}
    {parameter=\v,performance=\port,safety=\comp,
     switch=\perf,color=\clr}
    {
        \draw[\clr,line width=1.5pt,opacity=1]
            (-1*\A+90:\port/\datascale) -- 
            (-2*\A+90:\comp/\datascale) -- 
            (-3*\A+90:\perf/\datascale) -- cycle;
            %(\A*4:\use/\datascale) -- 
            %(\A*5:\secur/\datascale) -- 
            %(\A*6:\rel/\datascale) -- 
            %(\A*7:\func/\datascale) -- cycle;
       % add legend
       \draw [line width=1.5pt,\clr,opacity=1] (\R*0.53,\R-10pt*\thecsvrow) -- ++(1,0) node[right,opacity=1,black] {\v};
    }
\end{tikzpicture}
\caption{Sensitivity analysis radar chart. We collect data by fixing $\alpha_1$ and $\alpha_2$ and vary $\alpha_3$ from 0 to 1; then, we fix $\alpha_2$ and $\alpha_3$ and vary $\alpha_1$ from 0 to 1. The travel time (lower is better) is reformed as the performance score (higher is better) in the graph. For comprehensibility, we plot $6$ out of $25$ results with different colors and the rest of the other results in blue with low opacity. Generally, we do not consider tuning $\alpha_2$ as it can harm the system's safety.}
\label{fig:sensitivity}
\end{figure}
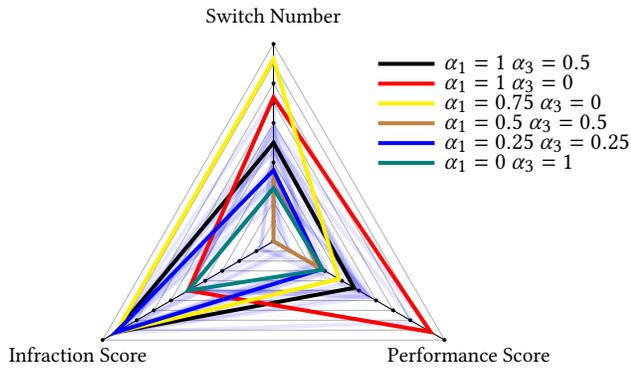
%\end{document}

%% file: graphs/intermittent_failure.tex
\usepgfplotslibrary{statistics}
\pgfplotsset{
/pgfplots/custom legend/.style={
legend image code/.code={
\draw [only marks,mark=square]
plot coordinates { 
(0.3cm,0cm)
};
}, },
}

\begin{figure*}
\begin{tikzpicture}
\begin{groupplot}[group style = {group size = 1 by 2, horizontal sep = 2cm, vertical sep = 0.75cm }]

\nextgroupplot[
        height=4cm, width=\textwidth,
      %bugsResolvedStyle/.style={},
      %xlabel={Track 1},
      ylabel={Travel Time (s)},
     ymin=0,
     xmin=0,
     ymax = 700,
     xmax = 6,
     xtick={0.75,2.25,3.75,5.25},
     xticklabels={{Track 1},{Track 2},{Track 3},{Track 4}},
     ymajorgrids,
     boxplot/draw direction=y,
     cycle list={{red},{blue}},
        %enlarge y limits=.2,
    %/pgfplots/boxplot/box extend=0.3,
      custom legend,
        legend style={at={(1,1)},anchor=south east,legend columns=6,
            column sep=0.5em}
]

% \pgfplotsset{
%     boxplot/draw/median/.code={
%         \draw [/pgfplots/boxplot/every median/.try]
%             (boxplot box cs:\pgfplotsboxplotvalue{median},0)
%             node[anchor=north west, font=\tiny] {\pgfmathprintnumber{\pgfplotsboxplotvalue{median}}}
%             (boxplot box cs:\pgfplotsboxplotvalue{median},1);
%     },
% }

\addlegendentry{LBC}
\addlegendentry{AP}
\addlegendentry{SA}
\addlegendentry{GS}
\addlegendentry{DS}

\node at (axis cs:0.25,-20) [anchor=south][black!30!green] {\scriptsize $(30)$};
\node at (axis cs:0.5,-15) [anchor=south][red] {\scriptsize {$\textbf{(0)}$}};
\node at (axis cs:0.75,-15) [anchor=south][brown] {\scriptsize {$\textbf{(0)}$}};
\node at (axis cs:1,-15) [anchor=south][black] {\scriptsize $(2)$};
\node at (axis cs:1.25,-15) [anchor=south][blue] {\scriptsize $\textbf{(0)}$};

%\node at (axis cs:1,282) [anchor=north] [black!30!green] {\tiny 273};
\node at (axis cs:0.5,600) [anchor=south] [red] {\tiny 253.45};
\node at (axis cs:0.75,600) [anchor=south] [brown] {\tiny 165.7};
\node at (axis cs:1,600) [anchor=south] [black] {\tiny 109.15};
\node at (axis cs:1.25,600) [anchor=south] [blue] {\tiny 160};    

\node at (axis cs:1.75,600) [anchor=south] [black!30!green] {\tiny 96.8};
\node at (axis cs:2.5,600) [anchor=south] [black] {\tiny 111.03};
\node at (axis cs:2.75,600) [anchor=south] [blue] {\tiny 112.63};

\node at (axis cs:3.25,600) [anchor=south] [black!30!green] {\tiny 265.22};
\node at (axis cs:3.5,600) [anchor=south] [red] {\tiny 607.05};
\node at (axis cs:3.75,600) [anchor=south] [brown] {\tiny 514.25};
\node at (axis cs:4,600) [anchor=south] [black] {\tiny 393.55};  
\node at (axis cs:4.25,600) [anchor=south] [blue] {\tiny 470.9};  

\node at (axis cs:4.75,600) [anchor=south] [black!30!green] {\tiny 84.85};
\node at (axis cs:5,600) [anchor=south] [red] {\tiny 185.8};
\node at (axis cs:5.25,600) [anchor=south] [brown] {\tiny 145.75};
\node at (axis cs:5.5,600) [anchor=south] [black] {\tiny 116.5};  
\node at (axis cs:5.75,600) [anchor=south] [blue] {\tiny 106};  
    
\addplot+ [black!30!green][boxplot={box extend=0.225, draw position=0.25},mark=x, mark=x] table [y index=0] {intermittent_failure_data/lbc/Track1_time.txt};

\addplot+ [red][boxplot={box extend=0.225, draw position=0.5},mark=x] table [y index=0] {intermittent_failure_data/ap/Track1_time.txt};
  % Ignore the miscalculations here
\addplot+ [brown][boxplot={box extend=0.225, draw position=0.75},mark=x] table [y index=0] {intermittent_failure_data/sa/Track1_time.txt};

\addplot+ [black][boxplot={box extend=0.225, draw position=1},mark=x] table [y index=0] {intermittent_failure_data/sag/Track1_time.txt};

\addplot+ [blue][boxplot={box extend=0.225, draw position=1.25},mark=x] table [y index=0] {intermittent_failure_data/dss/Track1_time.txt};
%%%%%%%%%%%%%%%%%%%%%%%%%%%%%%%%%%%%%%%%%%%%%%%%%%%%%%%%%%
\node at (axis cs:1.75,-15) [anchor=south][black!30!green] {\scriptsize $\textbf{(0)}$};
\node at (axis cs:2,-15) [anchor=south][red] {\scriptsize $(30)$};
\node at (axis cs:2.25,-15) [anchor=south][brown] {\scriptsize $(30)$};
\node at (axis cs:2.5,-15) [anchor=south][black] {\scriptsize {$(11)$}};
\node at (axis cs:2.75,-15) [anchor=south][blue] {\scriptsize {$(3)$}};

\addplot+ [black!30!green][boxplot={box extend=0.225, draw position=1.75}, mark=x] table [y index=0] {intermittent_failure_data/lbc/Track2_time.txt};

\addplot+ [red][boxplot={box extend=0.225, draw position=2},mark=x] table [y index=0] {intermittent_failure_data/ap/Track2_time.txt};

\addplot+ [brown][boxplot={box extend=0.225, draw position=2.25},mark=x] table [y index=0] {intermittent_failure_data/sa/Track2_time.txt};

\addplot+ [black][boxplot={box extend=0.225, draw position=2.5},mark=x] table [y index=0] {intermittent_failure_data/sag/Track2_time.txt};
\addplot+ [blue][boxplot={box extend=0.225, draw position=2.75},mark=x] table [y index=0] {intermittent_failure_data/dss/Track2_time.txt};    

%%%%%%%%%%%%%%%%%%%%%%%%%%%%%%%%%%%%%%%%%%%%%%%%%%%%%%%%%%%%%%%%%
\node at (axis cs:3.25,-15) [anchor=south][black!30!green] {\scriptsize $(3)$};
\node at (axis cs:3.5,-15) [anchor=south][red] {\scriptsize {$\textbf{(0)}$}};
\node at (axis cs:3.75,-15) [anchor=south][brown] {\scriptsize $\textbf{(0)}$};
\node at (axis cs:4.0,-15) [anchor=south][black] {\scriptsize $(8)$};
\node at (axis cs:4.25,-15) [anchor=south][blue] {\scriptsize {$\textbf{(0)}$}};

\addplot+ [black!30!green][boxplot={box extend=0.225, draw position=3.25},mark=x] table [y index=0] {intermittent_failure_data/lbc/Track3_time.txt};

\addplot+ [red][boxplot={box extend=0.225, draw position=3.5},mark=x] table [y index=0] {intermittent_failure_data/ap/Track3_time.txt};
  % Ignore the miscalculations here
\addplot+ [brown][boxplot={box extend=0.225, draw position=3.75},mark=x] table [y index=0] {intermittent_failure_data/sa/Track3_time.txt};

\addplot+ [black][boxplot={box extend=0.225, draw position=4.0},mark=x] table [y index=0] {intermittent_failure_data/sag/Track3_time.txt};

\addplot+ [blue][boxplot={box extend=0.225, draw position=4.25},mark=x] table [y index=0] {intermittent_failure_data/dss/Track3_time.txt};
%%%%%%%%%%%%%%%%%%%%%%%%%%%%%%%%%%%%%%%%%%%%%%%%%%%%%%%%%%%%%%%%
\node at (axis cs:4.75,-15) [anchor=south][black!30!green] {\scriptsize $\textbf{(1)}$};
\node at (axis cs:5,-15) [anchor=south][red] {\scriptsize {$\textbf{(0)}$}};
\node at (axis cs:5.25,-15) [anchor=south][brown] {\scriptsize {$\textbf{(0)}$}};
\node at (axis cs:5.5,-15) [anchor=south][black] {\scriptsize {$\textbf{(0)}$}};
\node at (axis cs:5.75,-15) [anchor=south][blue] {\scriptsize {$\textbf{(0)}$}};

\addplot+ [black!30!green][boxplot={box extend=0.225, draw position=4.75},mark=x] table [y index=0] {data/lbc/Track4_time.txt};
\addplot+ [boxplot={box extend=0.225, draw position=5},mark=x] table [y index=0] {data/ap/Track2_time.txt};
  % Ignore the miscalculations here
\addplot+ [brown][boxplot={box extend=0.225, draw position=5.25},mark=x] table [y index=0] {data/sa/Track2_time.txt};
\addplot+ [black][boxplot={box extend=0.225, draw position=5.5},mark=x] table [y index=0] {data/sag/Track4_time.txt};

\addplot+ [blue][boxplot={box extend=0.225, draw position=5.75},mark=x] table [y index=0] {data/dss/Track4_time.txt};

\draw [dashed] (1.5,0) -- (1.5,700);
\draw [dashed] (3,0) -- (3,700);
\draw [dashed] (4.5,0) -- (4.5,700);
]

\nextgroupplot[
        height=4cm, width=\textwidth,
      %bugsResolvedStyle/.style={},
      %xlabel={Track 1},
      ylabel={Infraction Score},
     ymin=0,
     xmin=0,
     ymax = 1.15,
     xmax = 6,
     xtick={0.75,2.25,3.75,5.25},
     xticklabels={{Track 1},{Track 2},{Track 3},{Track 4}},
     ymajorgrids,
     boxplot/draw direction=y,
     cycle list={{red},{blue}},
        %enlarge y limits=.2,
    %/pgfplots/boxplot/box extend=0.3,
      custom legend,
        legend style={at={(1,1)},anchor=south east,legend columns=6,
            column sep=0.5em}
]

%\pgfplotsset{
%     boxplot/draw/median/.code={
%         \draw [/pgfplots/boxplot/every median/.try]
%             (boxplot box cs:\pgfplotsboxplotvalue{median},0)
%             node[anchor=north west, font=\tiny] {\pgfmathprintnumber{\pgfplotsboxplotvalue{median}}}
%            (boxplot box cs:\pgfplotsboxplotvalue{median},1);
%     },
% }

\node at (axis cs:1,0) [anchor=south][black] {\scriptsize $\textbf{(4.1)}$};
\node at (axis cs:1.25,0) [anchor=south][blue] {\scriptsize $(4.5)$};

\node at (axis cs:2.5,0) [anchor=south][black] {\scriptsize $\textbf{(1.5)}$};
\node at (axis cs:2.75,0) [anchor=south][blue] {\scriptsize $(1.8)$};

\node at (axis cs:4,0) [anchor=south][black] {\scriptsize $\textbf{(4.0)}$};
\node at (axis cs:4.25,0) [anchor=south][blue] {\scriptsize $\textbf{(4.2)}$};

\node at (axis cs:5.5,0) [anchor=south][black] {\scriptsize $\textbf{(2.2)}$};
\node at (axis cs:5.75,0) [anchor=south][blue] {\scriptsize $(3.3)$};

\node at (axis cs:0.25,1) [anchor=south] [black!30!green] {\tiny 0.59};
\node at (axis cs:0.5,1) [anchor=south] [red] {\tiny 1};
\node at (axis cs:0.75,1) [anchor=south] [brown] {\tiny 1};
\node at (axis cs:1,1) [anchor=south] [black] {\tiny 0.75};
\node at (axis cs:1.25,1) [anchor=south] [blue] {\tiny 1};    

\node at (axis cs:1.75,1) [anchor=south] [black!30!green] {\tiny 1};
\node at (axis cs:2,1) [anchor=south] [red] {\tiny 0.58};
\node at (axis cs:2.25,1) [anchor=south] [brown] {\tiny 0.86};
\node at (axis cs:2.5,1) [anchor=south] [black] {\tiny 1};
\node at (axis cs:2.75,1) [anchor=south] [blue] {\tiny 1};

\node at (axis cs:3.25,1) [anchor=south] [black!30!green] {\tiny 0.75};
\node at (axis cs:3.5,1) [anchor=south] [red] {\tiny 1};
\node at (axis cs:3.75,1) [anchor=south] [brown] {\tiny 1};
\node at (axis cs:4,1) [anchor=south] [black] {\tiny 1};  
\node at (axis cs:4.25,1) [anchor=south] [blue] {\tiny 1};  

\node at (axis cs:4.75,1) [anchor=south] [black!30!green] {\tiny 1};
\node at (axis cs:5,1) [anchor=south] [red] {\tiny 1};
\node at (axis cs:5.25,1) [anchor=south] [brown] {\tiny 1};
\node at (axis cs:5.5,1) [anchor=south] [black] {\tiny 1};  
\node at (axis cs:5.75,1) [anchor=south] [blue] {\tiny 1};

\addplot+ [black!30!green][boxplot={box extend=0.225, draw position=0.25},mark=x, mark=x] table [y index=0] {intermittent_failure_data/lbc/Track1_safe.txt};

\addplot+ [red][boxplot={box extend=0.225, draw position=0.5},mark=x] table [y index=0] {intermittent_failure_data/ap/Track1_safe.txt};
  % Ignore the miscalculations here
\addplot+ [brown][boxplot={box extend=0.225, draw position=0.75},mark=x] table [y index=0] {intermittent_failure_data/sa/Track1_safe.txt};

\addplot+ [black][boxplot={box extend=0.225, draw position=1},mark=x] table [y index=0] {intermittent_failure_data/sag/Track1_safe.txt};

\addplot+ [blue][boxplot={box extend=0.225, draw position=1.25},mark=x] table [y index=0] {intermittent_failure_data/dss/Track1_safe.txt};
%%%%%%%%%%%%%%%%%%%%%%%%%%%%%%%%%%%%%%%%%%%%%%%%%%%%%%%%%%

\addplot+ [black!30!green][boxplot={box extend=0.225, draw position=1.75}, mark=x] table [y index=0] {intermittent_failure_data/lbc/Track2_safe.txt};

\addplot+ [red][boxplot={box extend=0.225, draw position=2},mark=x] table [y index=0] {intermittent_failure_data/ap/Track2_safe.txt};

\addplot+ [brown][boxplot={box extend=0.225, draw position=2.25},mark=x] table [y index=0] {intermittent_failure_data/sa/Track2_safe.txt};

\addplot+ [black][boxplot={box extend=0.225, draw position=2.5},mark=x] table [y index=0] {intermittent_failure_data/sag/Track2_safe.txt};
\addplot+ [blue][boxplot={box extend=0.225, draw position=2.75},mark=x] table [y index=0] {intermittent_failure_data/dss/Track2_safe.txt};    

%%%%%%%%%%%%%%%%%%%%%%%%%%%%%%%%%%%%%%%%%%%%%%%%%%%%%%%%%%%%%%%%%

\addplot+ [black!30!green][boxplot={box extend=0.225, draw position=3.25},mark=x] table [y index=0] {intermittent_failure_data/lbc/Track3_safe.txt};

\addplot+ [red][boxplot={box extend=0.225, draw position=3.5},mark=x] table [y index=0] {intermittent_failure_data/ap/Track3_safe.txt};
  % Ignore the miscalculations here
\addplot+ [brown][boxplot={box extend=0.225, draw position=3.75},mark=x] table [y index=0] {intermittent_failure_data/sa/Track3_safe.txt};

\addplot+ [black][boxplot={box extend=0.225, draw position=4.0},mark=x] table [y index=0] {intermittent_failure_data/sag/Track3_safe.txt};

\addplot+ [blue][boxplot={box extend=0.225, draw position=4.25},mark=x] table [y index=0] {intermittent_failure_data/dss/Track3_safe.txt};
%%%%%%%%%%%%%%%%%%%%%%%%%%%%%%%%%%%%%%%%%%%%%%%%%%%%%%%%%%%%%%%%

\addplot+ [black!30!green][boxplot={box extend=0.225, draw position=4.75},mark=x] table [y index=0] {intermittent_failure_data/lbc/Track4_safe.txt};
\addplot+ [red][boxplot={box extend=0.225, draw position=5},mark=x] table [y index=0] {intermittent_failure_data/ap/Track4_safe.txt};
  % Ignore the miscalculations here
\addplot+ [brown][boxplot={box extend=0.225, draw position=5.25},mark=x] table [y index=0] {intermittent_failure_data/sa/Track4_safe.txt};
\addplot+ [black][boxplot={box extend=0.225, draw position=5.5},mark=x] table [y index=0] {intermittent_failure_data/sag/Track4_safe.txt};

\addplot+ [blue][boxplot={box extend=0.225, draw position=5.75},mark=x] table [y index=0] {intermittent_failure_data/dss/Track4_safe.txt};

\draw [dashed] (1.5,0) -- (1.5,1.2);
\draw [dashed] (3,0) -- (3,1.2);
\draw [dashed] (4.5,0) -- (4.5,1.2);
]
\end{groupplot}
\end{tikzpicture}
\caption{Travel times and safety scores for the \ac{av} with intermittent camera occlusion on all 4 tracks. We show the number of times a controller failed to complete a track in parenthesis below each box, highlighting the least failures in bold. We also show the median of the distributions. $DS$ achieves the best performance among the controllers that do not fail. We also observe that $DS$ consistently has a better performance than $SA$ and less variance of infraction score than $GS$, indicating that $DS$ maintains a good balance between performance and safety. The mean of reverse switches performed by $GS$ and $DS$ across all tracks is also shown below each box in the graph of infraction score.}
\label{fig:interfailure}
\end{figure*}
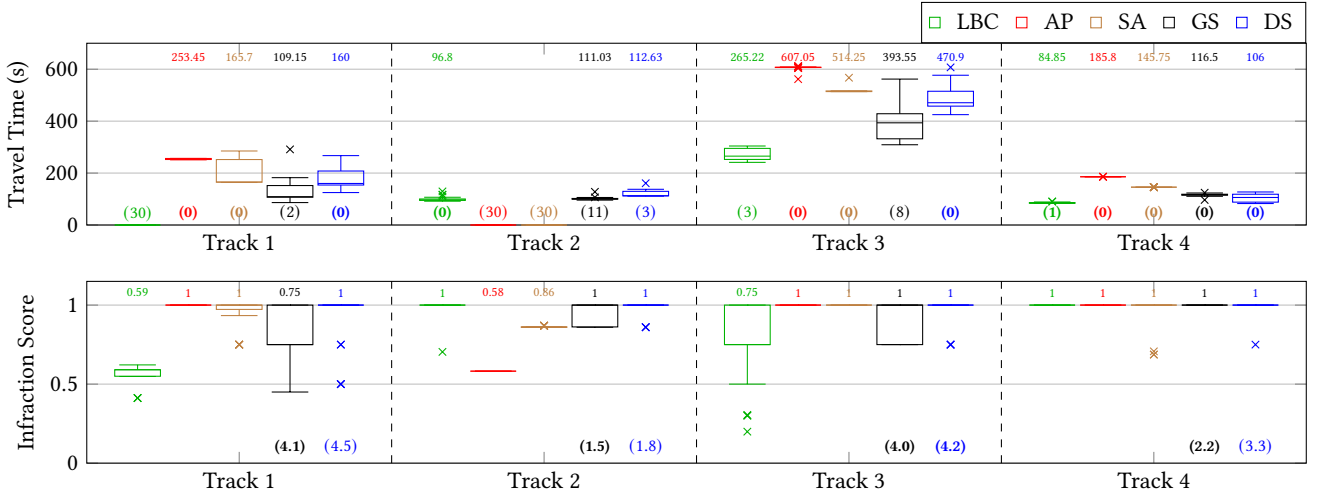

%the method works fairly well 

%% file: tables/computation_time.tex
\newcolumntype{P}[1]{>{\centering\arraybackslash}p{#1}}
\newcolumntype{M}[1]{>{\centering\arraybackslash}m{#1}}

\begin{table}[]
\caption {Computation Time of Non-Myopic Planner}
\normalsize
\centering
\renewcommand{\arraystretch}{1.5}
\resizebox{\columnwidth}{!}{
\begin{tabular}{|M{2cm}|M{2cm}|M{2cm}|M{2cm}|M{2cm}|}
\hline
MCTS Iterations & Average Computation Time (seconds) & Average Travel Time (seconds) & Average Infraction Score & Average Number of Switches \\ \hline
100             & \textbf{0.38}                                                                                  & 234.68                                                                      & 0.95                                                                & 6.25                                                                  \\ \hline
500             & 0.94                                                                                  & \textbf{196.84}                                                                      & \textbf{1}                                                                   & 5.42                                                                  \\ \hline
1000            & 1.64                                                                                  & 232.43                                                                      & \textbf{1}                                                                   & 5.00                                                                  \\ \hline
2000            & 3.00                                                                                  & 230.58                                                                      & \textbf{1}                                                                   & \textbf{3.33}                                                                  \\ \hline
\end{tabular}}
\label{tab:computation}
\end{table}

%% file: sections/related_work.tex
\section{Related Work}
\label{sec:rw}
% As discussed, Simplex architectures have long been used in the field of \ac{cps} to ensure the safety of \ac{cps} with unverifiable controllers. This has become more relevant with the increased use of \ac{ml} controllers to operate these systems. 
The conventional decision logic used in simplex architectures is based mainly on verification techniques such as linear matrix inequality~\cite{seto1999case} and reachability analysis~\cite{bak2011sandboxing}. For example, Johnson, Taylor T \emph{et al.}~\cite{johnson2016real} present a real-time reachability algorithm that uses the offline LMI results with online reachability analysis. A zero-level set of barrier certificates~\cite{prajna2004safety} that separates the unsafe region from all the possible system trajectories is presented as a decision logic for hybrid systems. While these approaches have worked well, they require an abstract system model, which is challenging to design for these complex autonomous systems with black-box \ac{ml} components. There have also been other non-verification approaches that do not require the abstract model. For example, Phan, Dung \emph{et al.}~\cite{DBLP:conf/acsd/PhanYCGSSS17} used a compositional proof technique called Assume-Guarantee contracts for switching between the controllers. A rule-based approach using historical performances of the controllers is designed for an unmanned aerial vehicle~\cite{vivekanandan2016simplex}. Such approaches are promising, they do not include the ability to perform the reverse switch, which is crucial for improving the system's performance. 

% There are relatively few thrusts addressing the reverse switch~\cite{vivekanandan2016simplex,desai2019soter,phan2020neural}. Jonhson \emph{et al.}~\cite{johnson2016real} propose the idea of reverse switching when the outputs of the performant controller are stabilized. In addition to the system's stability, reachability analysis has been used in several research findings~\cite{desai2019soter,DBLP:conf/nfm/MehmoodSBSS22} to check if the system is safe in the near future before making the reverse switch. Neural Simplex Architecture~\cite{phan2020neural} is a recent work that proposes two approaches to reverse switching. The first approach performs the switching by looking into the future through simulations. The second approach checks the system's current state in reference to the chosen safety boundary and decides if a switch is required. 
Although previous works have proposed various promising strategies for switching control to the safety controller, relatively little work has been done to investigate the decision logic for switching back to the performant controller~\cite{vivekanandan2016simplex,desai2019soter}. \citeauthor{desai2019soter}~\shortcite{desai2019soter} discuss a general framework for reverse switch that uses reachability analysis to check if the system is safe in the near future, irrespective of the controller that the system is running with. However, this decision logic results in a conservative reverse switch as the switch only happens when all reachable states of both controllers are in the safe set. Mehmood, Usama \emph{et al.}~\shortcite{DBLP:conf/nfm/MehmoodSBSS22} propose the Black-Box Simplex Architecture, which removes the requirement that the safety controller needs to be statically verified by burdening the decision module with extensive runtime checking. The reverse switch is driven by reachability analysis. Recently, many learning-based driving decision models have also been proposed~\cite{DBLP:conf/itsc/ChenYT19,DBLP:journals/jsa/RamakrishnaHBKD20,phan2020neural}. 
% Therefore, a series of works adapt simplex architecture to the learning-based controller to provide a safety guarantee for AI controllers. 
For example, the Neural Simplex Architecture~\cite{phan2020neural} is proposed to retrain and adapt the performant controller online and performs the reverse switch by leveraging reachability analysis.
% or if the current system state is far enough from the states which can trigger the forward switch. 
However, this architecture demands that both controllers are run in parallel and the neural network is updated online, significantly increasing the system's computation burden. 
Also, reachability analysis generally assumes an accurate system model is available; however, reachability
analysis for learning-based controllers is still limited to feed-forward ReLU-based networks~\cite{DBLP:journals/corr/LomuscioM17}.
% , and over-approximated reachable states can easily make overly conservative reverse switches. 
% Similarly, the approach proposed in~\cite{DBLP:journals/jsa/PengTSL22} retrains a graph attention-based reinforcement learning controller online.
On the other hand, our approach uses an \textit{anytime} algorithm \ac{mcts} to find a (near) optimal decision by exploring possible future trajectories in an asymmetric manner. Also, as the set of generative models used by the online heuristic search can be easily updated~\cite{mukhopadhyay2019online}, the proposed approach can seamlessly adapt to any exogenous changes in the environment.

%% file: sections/conclusion.tex
\section{Discussion}
\label{sec:discussion}
Finally, we conclude with a discussion about how safety and performance can be balanced in a complex cyber-physical system through dynamic switching between controllers. We hypothesize that the forward switch, which is critical for ensuring the safety of the system, must be: a) computationally cheap in terms of taking time to detect the threat to the system and making the decision for the switch; and b) must place utmost emphasis on the safety of the system. The reverse switch, on the other hand, has different constraints. First, it can afford relatively higher latency, i.e., the system can operate in the \textit{safe mode} while the decision for the reverse switch is computed. Second, the reverse switch must be non-myopic. Note that it is critical for the system to avoid facing the very same threats to safety that had caused the forward switch in the first place; as a result, careful consideration of the evolution of the system, conditional on the environmental parameters, must be performed to ensure that the reverse switch is safe. While the proposed data-driven approach in the paper is not verifiable, the Monte Carlo search is guaranteed to converge (given enough computational time) to the optimal action given the underlying Markov decision process. This observation further points out the need to accurately represent the Markovian process. An important consideration in the reverse switch is also the dynamic nature of the environment, i.e., the non-myopic decision-making approach must be equipped to consider any abrupt changes in the environmental parameters. This consideration is the major driver for the usage of an online search based approach in the proposed method, as opposed to a \textit{policy} trained offline (e.g., by using reinforced learning or dynamic programming) and then invoked instantaneously during execution.

\section{Conclusion}
\label{sec:conclusion}
We present the dynamic simplex strategy for controller selection in controller-redundant \ac{cps}. Our approach provides principled approaches for both forward and reverse switching. The approach balances safety and performance by leveraging a combination of a myopic action selector based on a surrogate model and an online non-myopic planner based on \ac{mcts} (for reverse switching). Our experimental evaluations using multiple \ac{av} examples in the CARLA simulator show that our approach with only a surrogate model can outperform other state-of-the-art alternatives in improving the system's performance without compromising safety under different environmental conditions and component failures. 

% Although our approach assumes safety verification protocols will be given to further guarantee safety, the open problem is that the existing verification techniques for neural network control systems mainly operate offline. Designing the techniques that can monitor systems' specifications online and augment the systems with the ability to avoid upcoming hazards is still a challenging problem~\cite{DBLP:journals/dt/TranXJ22}.}

% It has a logic that is both online and performs reverse switching. We model the logic as a sequential-decision problem and formulate it as a Semi-Markov decision process. The decision for a forward switch is performed by a myopic selector that uses a surrogate model to get the performant and safety controller's historical performance in a given state. The decision for a reverse switch is performed by a planner that simulates the future operating conditions using an online \ac{mcts}. Based on the switching decision, a switching routine is activated, which uses domain rules and the system's state to ensure a smooth transition between the controllers. Our experimental evaluations using an \ac{av} example in CARLA simulations have shown our approach to improve the system's performance while not compromising safety. 

% One potential direction of future work is to instantiate a chain of \ac{mcts} trees and average the scores across each tree to determine the optimal action. 

%% file: sections/appendix.tex
\section{Technical Appendix}
\input{tables/symbol_table}

\subsection{Controllers}\label{appendix:AV_setup}

\begin{figure}[hbt!]
 \centering
 \includegraphics[width=0.9\columnwidth]{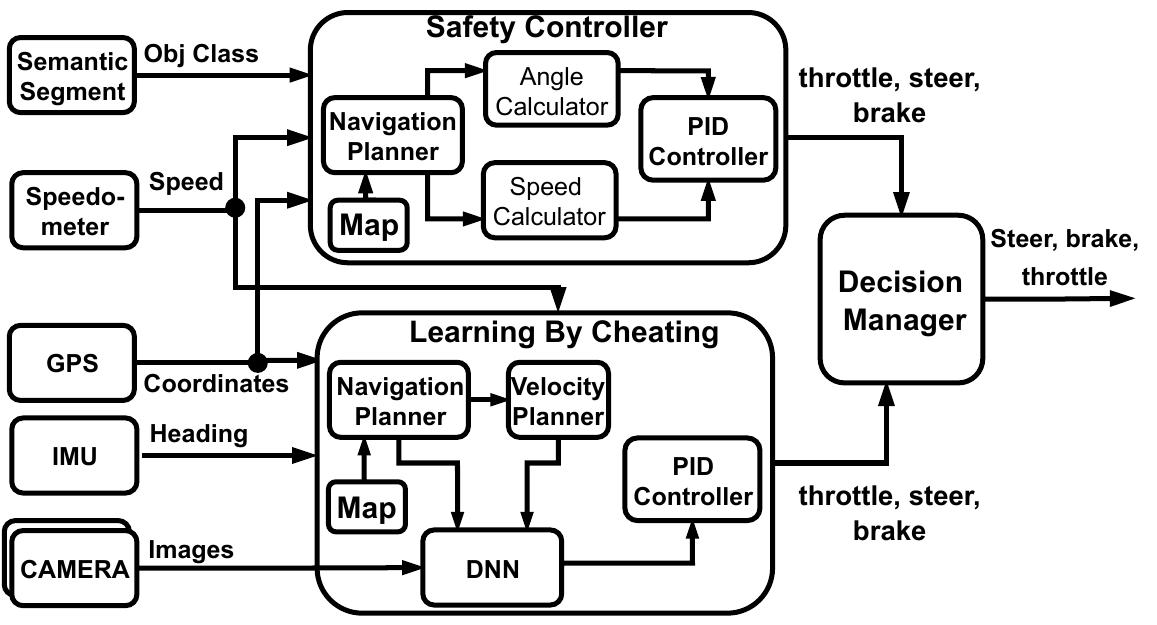}
 \caption{System model of the example \ac{av} in CARLA simulation.}
%  The \ac{av} is primarily driven by the \ac{lbc} controller~\cite{chen2020learning}. We have augmented a safety controller and a decision logic.}
%   which is supervised by an AEBS for emergency braking.} 
 \label{fig:sys-model}
\end{figure}

\textbf{Performant Contoller}: It uses a navigation planner that takes the waypoint information from the simulator and divides them into smaller position targets. Next, it uses the \ac{gps} and \ac{imu} sensors to get the vehicle's current position. It feeds this along with the position targets into a velocity planner that computes the desired speed. The desired speed and camera images are fed into a \ac{dnn}, which predicts the trajectory angle and the target speed. These predictions and the current speed is sent to PID controllers to compute the throttle, brake, and steer control signals. 

\textbf{Autopilot Controller}: The controller uses the \ac{gps}, the speedometer, and the semantic segmentation camera sensor to compute the control actions as follows. First, it uses the navigation planner to get the preset waypoints and divides them into smaller position targets using the priority information (e.g., position of traffic signs) from the simulator. The position targets, current position, and speed (from \ac{gps} and \ac{imu} sensors) are sent to angle and speed calculator functions to calculate the trajectory and the desired speed. These values are sent to different PID controllers to compute the throttle, brake, and steer control signals. Finally, the control actions from the two controllers are forwarded to a decision manager with the logic discussed in \cref{sec:approach}. 

\textbf{Controller Operation}: In addition to these rules, we also use a warm-up phase during which warms up the controller selected by the decision-maker before bringing it online to operate the system. The warm-up is required because we do not run both controllers in parallel to avoid computational costs and save energy, i.e., when one controller is operating, the other controller is idle and is unaware of the system's current state. If the decision-maker decides to switch, the routine starts to run the idle controller in shadow mode to slowly update it with the system's current state. After the warm-up phase is completed, the selected controller's actions are taken predictions start being used for operating the system.

\begin{figure}[hbt!]
 \centering
 \includegraphics[width=0.8\columnwidth]{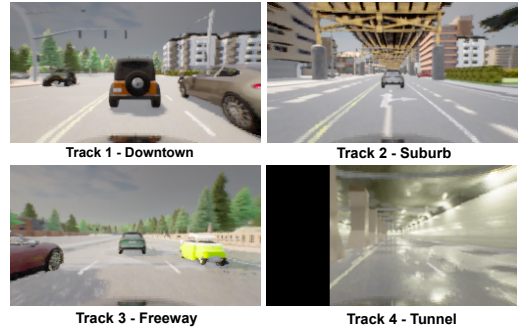}
 \caption{Samples of the tracks as captured by the center camera attached to the \ac{av}. The occlusion in the track 4 image is because of a camera failure.}
 \label{fig:regions}
\end{figure}

\textbf{Track Description}: Track 1 is around downtown with high traffic density and has several traffic signs in most scenes. Track 2 is around the suburb with an overpass and typically has a low traffic density. Track 3 is around a long freeway with low traffic density for all the scenes. Finally, Track 4 runs through a tunnel and then enters a city with traffic lights and medium traffic density.

\input{graphs/ablation_study.tex}

\subsection{Ablation Study}\label{appendix:ablation study}
To understand the effects of the non-myopic planner and the domain rules, we conducted an ablation study. We show the results in \cref{fig:ablation}. D-Myopic and ND-Myopic indicate the configurations that use a myopic selector with domain rules and a myopic selector without domain rules for reverse switches, respectively. D-Nonmyopic and ND-Nonmyopic indicate the configurations that use a non-myopic planner with domain rules and a non-myopic planner without domain rules for reverse switching, respectively. We observe that coupling domain rules with different reverse-switching configurations have a minor impact on the system's travel times and infraction scores on Track 1, Track 2, and Track 4. However, both D-Nonmyopic and ND-Nonmyopic achieve a smaller variance of infraction score on Track 1, shorter travel time on Track 4, and fewer reverse switches across all tracks than D-Myopic and ND-Myopic, demonstrating that reverse switching with nonmyopic planner plays an essential role in improving the system's performance and stability. We also observe that removing domain rules from reverse-switching configurations leads to more failures on Track 3. We hypothesize that this behavior is driven by Track 3 being a long freeway with a low traffic density; therefore, both the performant controller and the safety controller can operate with higher velocities than they can operate on the other three tracks, which are similar in terms of velocities achieved by the controllers. On Track 3, switching without domain rules (e.g., reducing speed for switching) can have a detrimental effect on the system's stability.

\subsection{Generative Model}\label{appendix:generative_model}
We use different distributions and an artificial neural network as our generative models. First, we sample temporal features and traffic density with random distribution during data collection; therefore, we model the transition of temporal features and traffic density as so. Second, we model the distribution of the permanent sensor failure as a Weibull distribution. We learn the distribution parameters by maximizing the likelihood of sensor failure data. Third, to model the intermittent sensor failure rate, we use an exponential growth function to simulate the likelihood of occlusion conditioned on weather and location, i.e., the sunnier or heavier the precipitation is, the more likely it is to cause a temporary occlusion. We also ensure that such failures depend on the state's structural features, e.g., sunny conditions or precipitation is unlikely to cause occlusion if the vehicle is operating in a tunnel. Finally, to model the duration and the arrival time of runtime monitor alarms, we aggregate historical alarm data and compute the average duration and inter-arrival time of these alarms conditioned on different temporal features, structural features, and traffic density. Then we use these historical data to train an artificial neural network as our generative model.

%% file: tables/symbol_table.tex
\begin{table}[hbt!]
\footnotesize
\centering
\renewcommand{\arraystretch}{1.2}
\resizebox{\columnwidth}{!}{
\begin{tabular}{|l|l|}
\hline
Notation                 & Description      
\\ \hline
$w_t$                 & \begin{tabular}[c]{@{}l@{}}A sequence of scenes characterized by transitions in structural \\ features $w^s_t$ or in temporal features $w^q_t$ at time step $t$\end{tabular}
\\ \hline
$s_t$                     & \begin{tabular}[c]{@{}l@{}}State tuple $< v, w^s_t, w^q_t, C_t, \phi_t, \omega_t >$ at time step $t$\end{tabular}  \\ \hline
$a$                     & The action $a$ which can be taken  \\ \hline
$v_t$                        & Velocity of the vehicle                                                                                                                                                                                   \\ \hline
$C_t$                     & The controller $C_t$ $\in$ $\{C^p, C^s\}$ driving the system at time step $t$                                                                                   \\ \hline
$d_t$                     & The traffic density at time step $t$                                                                                   \\ \hline
$\phi_t$   & The failure state of n components                                                                                                                                                                         \\ \hline
$\psi_t$  & The runtime monitor state at time step $t$  
                                                                                    \\ \hline
$\omega_t$ & A counter that keeps track the number of switches has performed                                                                                                                                           \\ \hline
$R(s,a)$   & \begin{tabular}[c]{@{}l@{}}The scoring function which returns the reward given \\ state $s$ and action $a$\end{tabular} 
\\ \hline    
$\lambda(s,a)$ & \begin{tabular}[c]{@{}l@{}}$\lambda$ $\in$ $\{\lambda^{p}, \lambda^{f}, \lambda^{c}\}$ which returns performance score, the safety \\ score, or the cost of switch.\end{tabular}
\\ \hline
$m_s$ & \begin{tabular}[c]{@{}l@{}}The maximum number of switches may happen during the \\ planning horizon.\end{tabular}
\\ \hline
$\tau^{q}$ & \begin{tabular}[c]{@{}l@{}}The discrete time period after which the parameter with \\ the highest frequency is queried\end{tabular}
\\ \hline
$M$ & \begin{tabular}[c]{@{}l@{}}A set of generative models.\end{tabular}
\\ \hline

$G$ & \begin{tabular}[c]{@{}l@{}}Surrogate model.\end{tabular}
\\ \hline
$t_e$ & \begin{tabular}[c]{@{}l@{}}The estimated time that the system may take to arrive at next \\ closest structural scene.\end{tabular}
\\ \hline
\end{tabular}
}
\end{table}

%% file: graphs/ablation_study.tex
\usepgfplotslibrary{statistics}
\pgfplotsset{
/pgfplots/custom legend/.style={
legend image code/.code={
\draw [only marks,mark=square]
plot coordinates { 
(0.3cm,0cm)
};
}, },
}

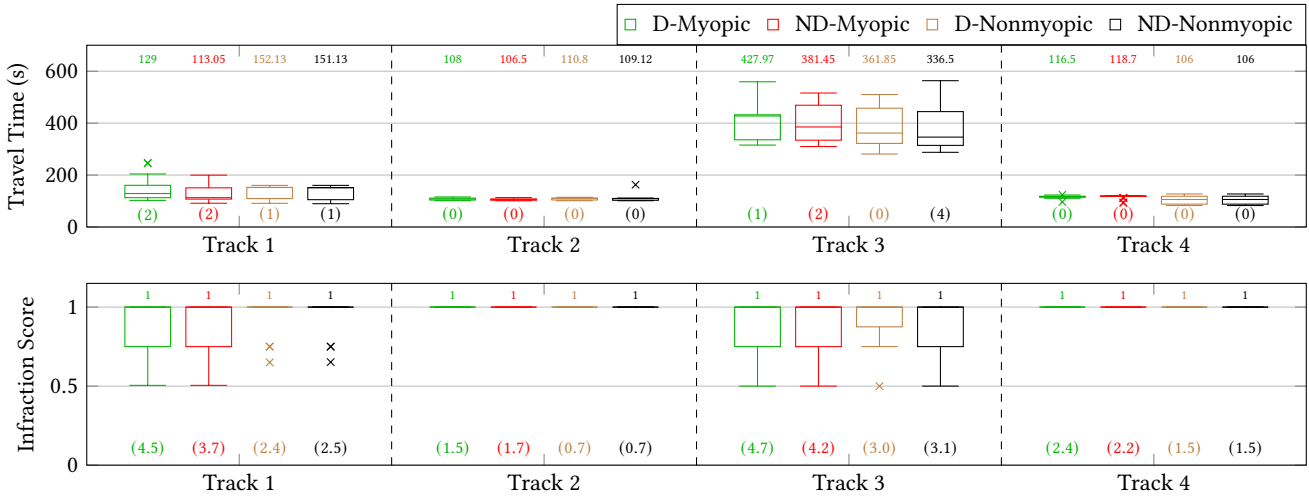
\begin{figure*}
\begin{tikzpicture}
\begin{groupplot}[group style = {group size = 1 by 2, horizontal sep = 2cm, vertical sep = 0.75cm }]

\nextgroupplot[
        height=4cm, width=\textwidth,
      %bugsResolvedStyle/.style={},
      %xlabel={Track 1},
      ylabel={Travel Time (s)},
     ymin=0,
     xmin=0,
     ymax = 700,
     xmax = 6,
     xtick={0.75,2.25,3.75,5.25},
     xticklabels={{Track 1},{Track 2},{Track 3},{Track 4}},
     ymajorgrids,
     boxplot/draw direction=y,
     cycle list={{red},{blue}},
        %enlarge y limits=.2,
    %/pgfplots/boxplot/box extend=0.3,
      custom legend,
        legend style={at={(1,1)},anchor=south east,legend columns=6,
            column sep=0.5em}
]

%\pgfplotsset{
%     boxplot/draw/median/.code={
%         \draw [/pgfplots/boxplot/every median/.try]
%             (boxplot box cs:\pgfplotsboxplotvalue{median},0)
%             node[anchor=north west, font=\tiny] {\pgfmathprintnumber{\pgfplotsboxplotvalue{median}}}
%             (boxplot box cs:\pgfplotsboxplotvalue{median},1);
%     },
% }

\addlegendentry{D-Myopic}
\addlegendentry{ND-Myopic}
\addlegendentry{D-Nonmyopic}
\addlegendentry{ND-Nonmyopic}

\node at (axis cs:0.3,-20) [anchor=south][black!30!green] {\scriptsize $(2)$};
\node at (axis cs:0.6,-15) [anchor=south][red] {\scriptsize {$(2)$}};
\node at (axis cs:0.9,-15) [anchor=south][brown] {\scriptsize {$(1)$}};
\node at (axis cs:1.2,-15) [anchor=south][black] {\scriptsize $(1)$};

%\node at (axis cs:1,282) [anchor=north] [black!30!green] {\tiny 273};
\node at (axis cs:0.3,600) [anchor=south] [black!30!green] {\tiny 129};
\node at (axis cs:0.6,600) [anchor=south] [red] {\tiny 113.05};
\node at (axis cs:0.9,600) [anchor=south] [brown] {\tiny 152.13};
\node at (axis cs:1.2,600) [anchor=south] [black] {\tiny 151.13};

\node at (axis cs:1.8,600) [anchor=south] [black!30!green] {\tiny 108};
\node at (axis cs:2.1,600) [anchor=south] [red] {\tiny 106.5};
\node at (axis cs:2.4,600) [anchor=south] [brown] {\tiny 110.8};
\node at (axis cs:2.7,600) [anchor=south] [black] {\tiny 109.12};

\node at (axis cs:3.3,600) [anchor=south] [black!30!green] {\tiny 427.97};
\node at (axis cs:3.6,600) [anchor=south] [red] {\tiny 381.45};  
\node at (axis cs:3.9,600) [anchor=south] [brown] {\tiny 361.85};
\node at (axis cs:4.2,600) [anchor=south] [black] {\tiny 336.5};

\node at (axis cs:4.8,600) [anchor=south] [black!30!green] {\tiny 116.5};
\node at (axis cs:5.1,600) [anchor=south] [red] {\tiny 118.7};  
\node at (axis cs:5.4,600) [anchor=south] [brown] {\tiny 106};  
\node at (axis cs:5.7,600) [anchor=south] [black] {\tiny 106};  

\addplot+ [black!30!green][boxplot={box extend=0.225, draw position=0.3},mark=x, mark=x] table [y index=0] {data/sag/Track1_time.txt};

\addplot+ [red][boxplot={box extend=0.225, draw position=0.6},mark=x] table [y index=0] {ablation/myopic/Track1_time.txt};
  % Ignore the miscalculations here
\addplot+ [brown][boxplot={box extend=0.225, draw position=0.9},mark=x] table [y index=0] {data/dss/Track1_time.txt};

\addplot+ [black][boxplot={box extend=0.225, draw position=1.2},mark=x] table [y index=0] {ablation/nonmyopic/Track1_time.txt};

%%%%%%%%%%%%%%%%%%%%%%%%%%%%%%%%%%%%%%%%%%%%%%%%%%%%%%%%%%
\node at (axis cs:1.8,-15) [anchor=south][black!30!green] {\scriptsize $(0)$};
\node at (axis cs:2.1,-15) [anchor=south][red] {\scriptsize $(0)$};
\node at (axis cs:2.4,-15) [anchor=south][brown] {\scriptsize $(0)$};
\node at (axis cs:2.7,-15) [anchor=south][black] {\scriptsize {$(0)$}};

\addplot+ [black!30!green][boxplot={box extend=0.225, draw position=1.8}, mark=x] table [y index=0] {data/sag/Track2_time.txt};

\addplot+ [red][boxplot={box extend=0.225, draw position=2.1},mark=x] table [y index=0] {ablation/myopic/Track2_time.txt};

\addplot+ [brown][boxplot={box extend=0.225, draw position=2.4},mark=x] table [y index=0] {data/dss/Track2_time.txt};

\addplot+ [black][boxplot={box extend=0.225, draw position=2.7},mark=x] table [y index=0] {ablation/nonmyopic/Track2_time.txt};   

%%%%%%%%%%%%%%%%%%%%%%%%%%%%%%%%%%%%%%%%%%%%%%%%%%%%%%%%%%%%%%%%%
\node at (axis cs:3.3,-15) [anchor=south][black!30!green] {\scriptsize $(1)$};
\node at (axis cs:3.6,-15) [anchor=south][red] {\scriptsize {$(2)$}};
\node at (axis cs:3.9,-15) [anchor=south][brown] {\scriptsize $(0)$};
\node at (axis cs:4.2,-15) [anchor=south][black] {\scriptsize $(4)$};

\addplot+ [black!30!green][boxplot={box extend=0.225, draw position=3.3},mark=x] table [y index=0] {data/sag/Track3_time.txt};

\addplot+ [red][boxplot={box extend=0.225, draw position=3.6},mark=x] table [y index=0] {ablation/myopic/Track3_time.txt};
  % Ignore the miscalculations here
\addplot+ [brown][boxplot={box extend=0.225, draw position=3.9},mark=x] table [y index=0] {data/dss/Track3_time.txt};

\addplot+ [black][boxplot={box extend=0.225, draw position=4.2},mark=x] table [y index=0] {ablation/nonmyopic/Track3_time.txt};

%%%%%%%%%%%%%%%%%%%%%%%%%%%%%%%%%%%%%%%%%%%%%%%%%%%%%%%%%%%%%%%%
\node at (axis cs:4.8,-15) [anchor=south][black!30!green] {\scriptsize $(0)$};
\node at (axis cs:5.1,-15) [anchor=south][red] {\scriptsize {$(0)$}};
\node at (axis cs:5.4,-15) [anchor=south][brown] {\scriptsize {$(0)$}};
\node at (axis cs:5.7,-15) [anchor=south][black] {\scriptsize {$(0)$}};

\addplot+ [black!30!green][boxplot={box extend=0.225, draw position=4.8},mark=x] table [y index=0] {data/sag/Track4_time.txt};
\addplot+ [red][boxplot={box extend=0.225, draw position=5.1},mark=x] table [y index=0] {ablation/myopic/Track4_time.txt};
  % Ignore the miscalculations here
\addplot+ [brown][boxplot={box extend=0.225, draw position=5.4},mark=x] table [y index=0] {data/dss/Track4_time.txt};
\addplot+ [black][boxplot={box extend=0.225, draw position=5.7},mark=x] table [y index=0] {ablation/nonmyopic/Track4_time.txt};

\draw [dashed] (1.5,0) -- (1.5,700);
\draw [dashed] (3,0) -- (3,700);
\draw [dashed] (4.5,0) -- (4.5,700);
]

\nextgroupplot[
        height=4cm, width=\textwidth,
      %bugsResolvedStyle/.style={},
      %xlabel={Track 1},
      ylabel={Infraction Score},
     ymin=0,
     xmin=0,
     ymax = 1.15,
     xmax = 6,
     xtick={0.75,2.25,3.75,5.25},
     xticklabels={{Track 1},{Track 2},{Track 3},{Track 4}},
     ymajorgrids,
     boxplot/draw direction=y,
     cycle list={{red},{blue}},
        %enlarge y limits=.2,
    %/pgfplots/boxplot/box extend=0.3,
      custom legend,
        legend style={at={(1,1)},anchor=south east,legend columns=6,
            column sep=0.5em}
]

%\pgfplotsset{
%     boxplot/draw/median/.code={
%         \draw [/pgfplots/boxplot/every median/.try]
%             (boxplot box cs:\pgfplotsboxplotvalue{median},0)
%             node[anchor=north west, font=\tiny] {\pgfmathprintnumber{\pgfplotsboxplotvalue{median}}}
%             (boxplot box cs:\pgfplotsboxplotvalue{median},1);
%     },
% }

\node at (axis cs:0.3,0) [anchor=south][black!30!green] {\scriptsize $(4.5)$};
\node at (axis cs:0.6,0) [anchor=south][red] {\scriptsize $(3.7)$};
\node at (axis cs:0.9,0) [anchor=south][brown] {\scriptsize $(2.4)$};
\node at (axis cs:1.2,0) [anchor=south][black] {\scriptsize $(2.5)$};

\node at (axis cs:1.8,0) [anchor=south][black!30!green] {\scriptsize $(1.5)$};
\node at (axis cs:2.1,0) [anchor=south][red] {\scriptsize $(1.7)$};
\node at (axis cs:2.4,0) [anchor=south][brown] {\scriptsize $(0.7)$};
\node at (axis cs:2.7,0) [anchor=south][black] {\scriptsize $(0.7)$};

\node at (axis cs:3.3,0) [anchor=south][black!30!green] {\scriptsize $(4.7)$};
\node at (axis cs:3.6,0) [anchor=south][red] {\scriptsize $(4.2)$};
\node at (axis cs:3.9,0) [anchor=south][brown] {\scriptsize $(3.0)$};
\node at (axis cs:4.2,0) [anchor=south][black] {\scriptsize $(3.1)$};

\node at (axis cs:4.8,0) [anchor=south][black!30!green] {\scriptsize $(2.4)$};
\node at (axis cs:5.1,0) [anchor=south][red] {\scriptsize $(2.2)$};
\node at (axis cs:5.4,0) [anchor=south][brown] {\scriptsize $(1.5)$};
\node at (axis cs:5.7,0) [anchor=south][black] {\scriptsize $(1.5)$};

\node at (axis cs:0.3,1) [anchor=south] [black!30!green] {\tiny 1};
\node at (axis cs:0.6,1) [anchor=south] [red] {\tiny 1};
\node at (axis cs:0.9,1) [anchor=south] [brown] {\tiny 1};
\node at (axis cs:1.2,1) [anchor=south] [black] {\tiny 1};

\node at (axis cs:1.8,1) [anchor=south] [black!30!green] {\tiny 1};
\node at (axis cs:2.1,1) [anchor=south] [red] {\tiny 1};
\node at (axis cs:2.4,1) [anchor=south] [brown] {\tiny 1};
\node at (axis cs:2.7,1) [anchor=south] [black] {\tiny 1};

\node at (axis cs:3.3,1) [anchor=south] [black!30!green] {\tiny 1};
\node at (axis cs:3.6,1) [anchor=south] [red] {\tiny 1};
\node at (axis cs:3.9,1) [anchor=south] [brown] {\tiny 1};
\node at (axis cs:4.2,1) [anchor=south] [black] {\tiny 1};  

\node at (axis cs:4.8,1) [anchor=south] [black!30!green] {\tiny 1};
\node at (axis cs:5.1,1) [anchor=south] [red] {\tiny 1};
\node at (axis cs:5.4,1) [anchor=south] [brown] {\tiny 1};
\node at (axis cs:5.7,1) [anchor=south] [black] {\tiny 1};

\addplot+ [black!30!green][boxplot={box extend=0.225, draw position=0.3},mark=x, mark=x] table [y index=0] {data/sag/Track1_safe.txt};

\addplot+ [red][boxplot={box extend=0.225, draw position=0.6},mark=x] table [y index=0] {ablation/myopic/Track1_safe.txt};
  % Ignore the miscalculations here
\addplot+ [brown][boxplot={box extend=0.225, draw position=0.9},mark=x] table [y index=0] {data/dss/Track1_safe.txt};

\addplot+ [black][boxplot={box extend=0.225, draw position=1.2},mark=x] table [y index=0] {ablation/nonmyopic/Track1_safe.txt};

%%%%%%%%%%%%%%%%%%%%%%%%%%%%%%%%%%%%%%%%%%%%%%%%%%%%%%%%%%

\addplot+ [black!30!green][boxplot={box extend=0.225, draw position=1.8}, mark=x] table [y index=0] {data/sag/Track2_safe.txt};

\addplot+ [red][boxplot={box extend=0.225, draw position=2.1},mark=x] table [y index=0] {ablation/myopic/Track2_safe.txt};

\addplot+ [brown][boxplot={box extend=0.225, draw position=2.4},mark=x] table [y index=0] {data/dss/Track2_safe.txt};

\addplot+ [black][boxplot={box extend=0.225, draw position=2.7},mark=x] table [y index=0] {ablation/nonmyopic/Track2_safe.txt};

%%%%%%%%%%%%%%%%%%%%%%%%%%%%%%%%%%%%%%%%%%%%%%%%%%%%%%%%%%%%%%%%%

\addplot+ [black!30!green][boxplot={box extend=0.225, draw position=3.3},mark=x] table [y index=0] {data/sag/Track3_safe.txt};

\addplot+ [red][boxplot={box extend=0.225, draw position=3.6},mark=x] table [y index=0] {ablation/myopic/Track3_safe.txt};
  % Ignore the miscalculations here
\addplot+ [brown][boxplot={box extend=0.225, draw position=3.9},mark=x] table [y index=0] {data/dss/Track3_safe.txt};

\addplot+ [black][boxplot={box extend=0.225, draw position=4.2},mark=x] table [y index=0] {ablation/nonmyopic/Track3_safe.txt};

%%%%%%%%%%%%%%%%%%%%%%%%%%%%%%%%%%%%%%%%%%%%%%%%%%%%%%%%%%%%%%%%

\addplot+ [black!30!green][boxplot={box extend=0.225, draw position=4.8},mark=x] table [y index=0] {data/sag/Track4_safe.txt};
\addplot+ [red][boxplot={box extend=0.225, draw position=5.1},mark=x] table [y index=0] {ablation/myopic/Track4_safe.txt};
  % Ignore the miscalculations here
\addplot+ [brown][boxplot={box extend=0.225, draw position=5.4},mark=x] table [y index=0] {data/dss/Track4_safe.txt};
\addplot+ [black][boxplot={box extend=0.225, draw position=5.7},mark=x] table [y index=0] {ablation/nonmyopic/Track4_safe.txt};

\draw [dashed] (1.5,0) -- (1.5,1.2);
\draw [dashed] (3,0) -- (3,1.2);
\draw [dashed] (4.5,0) -- (4.5,1.2);
]
\end{groupplot}
\end{tikzpicture}
\caption{\textbf{Top:} Travel times of the different controller configurations across the $4$ tracks \textbf{(lower is better)}. We show the number of times a controller failed to complete a track in parenthesis below each box. We also show the median of the distributions at the top of the graph. \textbf{(Bottom):} We show the infraction score \textbf{(higher is better)} for all controller configurations across all tracks. We also show the mean of reverse switches performed by different configurations below each box.}
\label{fig:ablation}
\end{figure*}

%the method works fairly well 